\def\doublecolumn{1} 
\if 1\doublecolumn
  \documentclass[10pt,journal]{IEEEtran}
\else
  \documentclass[12pt, journal, onecolumn]{IEEEtran}
\fi

\def\blind{1} 

\usepackage{mypackage}

\newcommand{\AirGCCD}{{\textsf{AirGC-CD}}\xspace}
\newcommand{\snr}{\mathtt{SNR}}
\newcommand{\clip}{\mathtt{clip}}
\newcommand{\FFT}{\mathtt{FFT}}
\newcommand{\IFFT}{\mathtt{IFFT}}
\newcommand{\PAPR}{\mathtt{PAPR}}

\newcommand{\Esk}{\mathcal{E}_{\mathrm{sk}}}
\newcommand{\Ecl}{\mathcal{E}_{\mathrm{cl}}}
\newcommand{\Ech}{\mathcal{E}_{\mathrm{ch}}}

\begin{document}

\title{
    \textsf{AirGC-CD}: Gaussian-Circulant Precoding for Exactly Debiasable PAPR Reduction in Over-the-Air Federated Learning
}

\if 1\blind
\author{Jonggyu Jang,~\IEEEmembership{Member,~IEEE}, Hyeonsu Lyu,~\IEEEmembership{Member,~IEEE}, and Hyun Jong Yang,~\IEEEmembership{Senior Member,~IEEE}
    \thanks{
    J. Jang is with the Department of Electronics Engineering, Chungnam National University, Daejeon 34134, Republic of Korea (e-mail: jgjang@cnu.ac.kr).
    H. Lyu and H. J. Yang are with the Institute of New Media and Communications, Seoul 08826, Republic of Korea (e-mail: \{hs.lyu, hjyang\}@snu.ac.kr).
    H. J. Yang is also with the Department of Electrical and Computer Engineering, Seoul National University, Seoul 08826, Republic of Korea.
    The corresponding author is Hyun Jong Yang.
    }
}
\else
\author{Anonymous Submission
    }
\fi
\maketitle

\begin{abstract}
Over-the-air federated learning lets edge devices transmit their local updates simultaneously, reducing the communication overhead.
The resulting waveform, however, has a peak-to-average power ratio (PAPR) that grows with the model dimension, and keeping the amplifier in its linear range leaves two remedies: clipping the peaks or backing off the transmit power.
Neither remedy is without cost: i) the clipping distortion appears at the receiver as a bias that cannot be removed, and ii) back-off keeps the signal intact but degrades the average signal-to-noise ratio (SNR).
Independent of this trade-off, the transmission remains uncompressed, spending one channel use per model parameter, which keeps large-model training out of reach.
To address these challenges, we propose \AirGCCD, an over-the-air scheme that precodes each local update with a partial Gaussian circulant matrix before clipping.
In \AirGCCD, the precoder's output is \textit{exactly} Gaussian regardless of the update's sparsity, so the clipping function is designed for a known distribution instead of inheriting it from the data.
This enables the clipping to be inverted on average by a single scalar Bussgang gain in closed form, and we prove that the resulting aggregate is \textit{exactly unbiased}, with clipping adding only variance. 
The clipping ratio is then the only free parameter left, trading the variance of the clipping against the SNR loss from back-off, and we derive its near-optimum in closed form.
Since the precoder is linear, it also acts as a compressor, reducing the transmission from the model dimension $d$ to the sketch dimension $m$ at a cost of only $\mathcal{O}(d\log d)$ via two fast Fourier transforms, whereas a Gaussian sketch costs $\mathcal{O}(md)$.
We then provide a convergence analysis under peak power constraints, yielding an $\mathcal{O}(1/\sqrt{T})$ convergence bound without a \textit {bias floor}. 
Experiments on five image datasets show that \AirGCCD outperforms baseline over-the-air FL schemes in most settings, particularly at low SNR, while using fewer channel uses per round.
\end{abstract}

\section{Introduction}

\IEEEPARstart{M}{achine} learning (ML) has been widely adopted across fields as diverse as computer vision, language processing, network optimization, and medical diagnosis, and is expected to be native to sixth-generation (6G) networks~\cite{brinton2024key,ruzomberka2023challenges}.
The data that matters most, however, is produced at the network edge, e.g., by phones, vehicles, sensors, and hospital servers, where it is often private or simply too voluminous to upload.
For training models on data that cannot be shared, federated learning (FL) has been a promising solution~\cite{pmlr-v54-mcmahan17a}, in which the edge devices (EDs) exchange only local model updates, and the parameter server (PS) aggregates those updates and broadcasts the result.

In the early days of ML, exchanging updates as large as the model size was unproblematic, since models were small relative to the datasets used to train them.
However, this is no longer the case: model sizes have grown tenfold per year~\cite{wu2023peta}, the per-round uplink payload is now comparable to the data itself, and communication has become the principal bottleneck of FL.
The problem is aggravated under orthogonal multiple access (OMA), where the uplink cost grows not only with the model size $d$ but also with the number of EDs, each requiring its own resource block.

To remove the dependence on the number of EDs, several existing studies adopt over-the-air computation (AirComp) for FL~\cite{yang2020federated,8870236,8849334}.
Rather than separating the devices in time or frequency, AirComp lets all EDs transmit analog-modulated updates simultaneously and exploits the superposition of the multiple-access channel to compute a nomographic function of their signals during transmission.
The aggregation required by FL is precisely such a function, so the channel performs it as part of the transmission, and a single round consumes the same resources, regardless of how many devices participate.

In analog AirComp-FL, however, \emph{the transmit amplitudes are the update values themselves}, which causes two problems that stem from the learning signal rather than from the communication channel.
First, one channel use is spent per model parameter, so a round still occupies $d$ symbols.
Second, the waveform inherits the statistics of deep-network updates, which are heavy-tailed and often nearly sparse, so the instantaneous transmit power exceeds the linear range of the radio-frequency power amplifier (PA) far more frequently than that of a digital constellation.
This was quantified recently: the authors of~\cite{bielefeld2026signalpeakpowerconstraint} measured single-carrier AirComp-FL waveforms with peak-to-average power ratio (PAPR) exceeding 30 dB, under which conventional iterative clipping and filtering (ICF)~\cite{armstrong2002peak, 975762} degrades convergence of learning.

\begin{table*}[t]
\centering
\caption{Positioning of \AirGCCD{} against representative AirComp-FL and PAPR-reduction schemes.}
\label{tab:related}
\footnotesize\setlength{\tabcolsep}{4pt}
\begin{tabular}{@{}lcccccc@{}}
\toprule
Scheme & Compression & Superposition & PAPR reduction & Unbiased aggregation & Encoding complexity & Year \\
\midrule
AirComp-FL~\cite{yang2020federated,9076343} & \xmarkg & \cmark & \xmarkg & \cmark & $\mathcal{O}(d)$ & 2020 \\
Sparsification + CS~\cite{8849334} & \cmark & \cmark & \xmarkg & \xmarkg & $\mathcal{O}(md)$ & 2019 \\
One-bit aggregation~\cite{9272666} & \cmark & \cmark & \cmark & \xmarkg & $\mathcal{O}(d)$ & 2021 \\
Power control / scheduling~\cite{9606731} & \xmarkg & \cmark & \xmarkg & \cmark & $\mathcal{O}(d)$ & 2022 \\
$1$-bit CS AirComp~\cite{9912341} & \cmark & \cmark & \xmarkg & \xmarkg & $\mathcal{O}(md)$ & 2023 \\
Sparse one-bit + error feedback~\cite{oh2024communication} & \cmark & \cmark & \xmarkg & \xmarkg & $\mathcal{O}(d)$ & 2024 \\
Sketching~\cite{jang2024fed,10472332} & \cmark & \cmark & \xmarkg & \cmark & $\mathcal{O}(md)$ or $\mathcal{O}(d\log d)$ & 2024 \\
ICF for AirComp-FL~\cite{bielefeld2026signalpeakpowerconstraint} & \xmarkg & \cmark & \cmark & \xmarkg & $\mathcal{O}(d\log d)$ & 2026 \\
\midrule
\textbf{\AirGCCD{} (ours)} & \cmark & \cmark & \cmark & \cmark & $\mathcal{O}(d\log d)$ & --- \\
\bottomrule
\end{tabular}
\end{table*}

\subsection{Related Work and Research Challenges}

Subsequent studies have extended AirComp-FL in two directions.
On the transceiver side, uniform-forcing designs~\cite{8364613} equalize the channel so that the receiver sum is unbiased, and joint device selection and power control~\cite{9606731} maximize the number of participating EDs under a transmit power budget; adaptive aggregation weights remove the need for transmitter-side CSI~\cite{azimi2024over}; and retransmissions trade channel uses for a lower aggregation error~\cite{hellstrom2023federated}.
Orthogonal to the transmission design, the number of communication rounds itself has been reduced by second-order updates~\cite{yang2022over}.
On the compression side, gradient sparsification~\cite{8849334,zhong2025over}, one-bit quantization~\cite{9272666}, quantization~\cite{qiao2024massive,xie2026joint,li2026map}, compressed sensing~\cite{9912341}, and randomized sketching~\cite{jang2024fed} have been studied.
Several of these compressors are biased, e.g., sparsification and one-bit quantization; the standard remedy is error feedback: the information a device discards is stored and added to its next update~\cite{oh2024communication}.
That remedy is unavailable here, since the accumulated residual can amplify the next update and thereby raise the PAPR.
In both directions, the transmit power is constrained only in the \textit{average sense}.
In~\cref{tab:related}, we summarize the related studies in terms of i) PAPR reduction for unbiased aggregation and ii) clipping-aware compression.

\paragraph*{C1) PAPR reduction for unbiased aggregation}
Under the peak-power constraint, the clipping level introduces a tradeoff between \textit{average power} and \textit{signal distortion}.
If an ED clips at a low level, the PAPR of the transmitted block decreases; hence, power control can raise the average transmit power toward the peak budget, which increases the received SNR and reduces the channel-noise term of the aggregation MSE.
However, a lower clipping level distorts the signal more severely, and the distortion arrives at the receiver as a bias that cannot be removed by averaging over time.
Conversely, if the ED clips at a high level or transmits without clipping, the signal remains intact, but the transmitter must back off by the largest element of the update; thus, the average transmit power decreases and the aggregation MSE increases.
As shown in~\cite{bielefeld2026signalpeakpowerconstraint}, the clipping level substantially influences the convergence of AirComp-FL, which motivates controlling the above tradeoff through the clipping level.
For raw local updates in FL, however, the distortion produced by a given clipping level depends on the training dataset and the current model weights, which makes the analysis \textit{intractable}; a PAPR reduction method that renders this tradeoff both analyzable and unbiased is therefore required.
Several solutions have been proposed in the PAPR reduction literature for PA, such as i) selective mapping (SLM)~\cite{bauml1996reducing}, ii) partial transmit sequences (PTS)~\cite{muller1997ofdm}, iii) ICF~\cite{armstrong2002peak,975762}, and iv) precoding~\cite{park2000papr}, yet none of them directly meets the above requirement.
In AirComp-FL, SLM and PTS let each device select among candidate rotations, which is incompatible with AirComp: the channel returns the desired average only if all devices apply the same map.
On the other hand, ICF preserves the superposition property and makes PAPR reduction analyzable by bounding the PAPR deterministically; nevertheless, it has an irremovable bias in gradient aggregation because the clipping biases the raw updates~\cite{bielefeld2026signalpeakpowerconstraint}.
Precoding can be directly applied to AirComp-FL, since a unitary transform shared by all devices leaves the superposition intact while spreading each coordinate's energy over the block, and Hadamard precoding~\cite{park2000papr} is its fast instance.
Its guarantees, however, are stated for a \textit{known modulation alphabet}, and Hadamard precoding cannot actively control the PAPR of the transmitted symbols; that is, the tradeoff between clipping and back-off cannot be addressed.

\paragraph*{C2) Clipping-aware compressor}
Independently of PAPR control, analog AirComp spends one channel use per model coordinate, so a round occupies $d$ symbols and compression is required for large models.
Compressors for AirComp-FL have been studied, including sparsification~\cite{8849334,zhong2025over} and one-bit compressed sensing~\cite{9912341,9272666}; however, these compressors are biased by themselves before any clipping is applied, and the standard remedy, error feedback~\cite{oh2024communication}, is unavailable here, since the accumulated residual amplifies the next update and raises the peak, which hinders convergence.
An alternative approach is Gaussian sketching~\cite{jang2024fed}, which is unbiased; its drawback is the cost of the dense matrix, $\mathcal{O}(md)$ complexity with a sketching dimension of $m$, which is prohibitive at $d\sim10^{7}$ and $m\sim10^4$.
The second challenge is to guarantee unbiasedness when the clipping operator is applied on top of the compressor, while overcoming the \textit{high complexity} of the Gaussian sketch.

\paragraph*{Research question}
Motivated by the above challenges, we identify the following research question:
\begin{mdframed}[outerlinecolor=black,outerlinewidth=1pt,linecolor=cccolor,middlelinewidth=1pt,roundcorner=0pt]
  \begin{center}
    \textbf{
    \textit{RQ: How can an ED find the best clipping strategy for accelerating the convergence, while compressing the transmission, and still let the server recover an \emph{exactly} unbiased aggregate at near-linear cost?}
    }
  \end{center}
\end{mdframed}

\subsection{Contributions}

To address the research question, we propose \AirGCCD, an over-the-air \textit{Gaussian-circulant} precoder with a \textit{clip-and-debias} policy. 
The design starts from precoding~\cite{park2000papr}, a classical PAPR-reduction mechanism that preserves superposition.
A classical precoder is chosen to \textit{flatten} the block, and its guarantee is then a statement about the resulting peak for a known alphabet.
We instead choose the precoder to obtain a \textit{known distribution} of transmitted symbols by i) randomizing the precoder with a seed broadcast to all EDs, and ii) subsampling the precoded block, which turns the same operator into a compressor. 
In particular, a circulant matrix generated by a Gaussian vector makes the transmitted symbols \textit{exactly Gaussian} with a common variance, at $\mathcal{O}(d\log d)$ complexity. 
By exploiting this \textit{known} distribution of the precoding output, we make the aggregated update exactly unbiased.
The detailed contributions are summarized as follows:
\begin{itemize}
    \item \textbf{Debiasing decoder:} We design a transceiver in which clipping remains \textit{debiasable}. Because the Gaussian-circulant precoder makes the precoded symbols Gaussian, the distortion introduced by clipping is inverted in the mean by a single Bussgang gain~\cite{bussgang1952crosscorrelation}, which depends only on the clipping ratio and is identical across EDs without any knowledge of updates.
    \item \textbf{Bias analysis:} We analyze a bias of clipped over-the-air aggregation, which has not been studied in either the AirComp or the PAPR literature. We prove that the proposed scheme is \textit{exactly} unbiased for every update and every clipping ratio with $\mathcal{O}(d\log d)$ complexity.
    \item \textbf{Convergence:} We analyze the convergence of the resulting scheme and show that the learning reaches a stationary point at the rate $\mathcal{O}(1/\sqrt{T})$, with \textit{no bias floor}. 
    \item \textbf{Optimal clipping level:} The clipping level is a dial between the two classical remedies, clipping and back-off. We derive the optimal clipping level for fast convergence.
\end{itemize}

\begin{figure*}
    \centering
    \includegraphics[width=\linewidth]{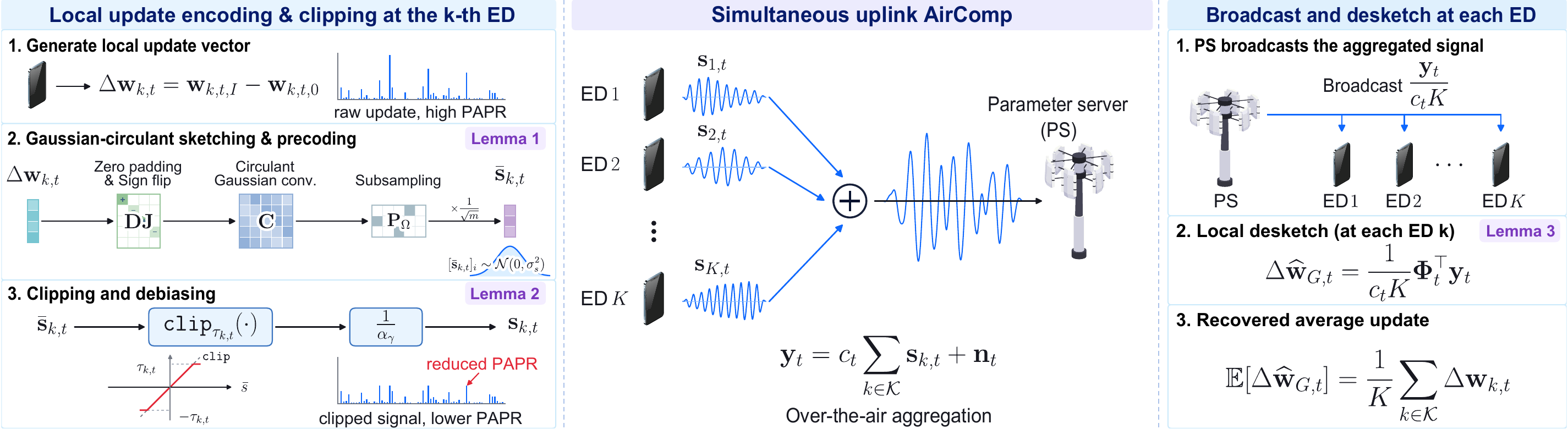}
    \caption{Schematic diagram of \AirGCCD. At each ED, the local model weight $\bw_{k,t,i}$ is trained on the local dataset via SGD. The EDs then encode their local update $\Delta\bw_{k,t}$ via Gaussian-circulant sketching. Next, to adjust the MSE tradeoff between the channel noise and clipping loss, we apply clipping, where $\alpha_\gamma$ denotes the Bussgang gain. In the AirComp uplink phase, the EDs simultaneously transmit $\bs_{k,t}$ over the shared channel. Then, the PS broadcasts the aggregated sketch to EDs via a reliable digital link. Finally, the EDs desketch and recover the average update. 
    }
    \label{fig:system_model}
\end{figure*}

\subsection{Notations}

Boldface lowercase and uppercase letters denote vectors and matrices, respectively. 
A scalar subscript indexes an element: $[\bg]_{j}$ is the $j$-th element of $\bg$; a set subscript denotes a subvector, i.e., the subvector of $\mathbf{u}$ indexed by $\mathcal S$ is $[\mathbf{u}]_{\mathcal S}$. 
We extend this indexing rule to matrices.
We write $\mathbf{I}_{d}$ for the $d\times d$ identity matrix, $(\cdot)^{\top}$ for the transpose, and $\odot$ for the elementwise (Hadamard) product. 
The clipping operator is $\clip_{\tau}(x):=\max(-\tau,\min(\tau,x))$, applied elementwise to vectors. 
For the standard normal distribution, $\phi(\cdot)$ and $Q(\cdot)$ denote the probability density function (PDF) and complementary cumulative distribution function (CCDF), respectively, i.e.,  $Q(x)=\int_{x}^\infty\phi(z)dz$.
For an integer $K$, $[K]$ represents the set of integers from $1$ to $K$, \ie, $[K] = \{1, 2, \dots, K\}$.

\section{System Model}\label{sec:system_model}

We consider a single-cell network in which $K$ EDs are served by a PS, co-located with the access point~\cite{yang2020federated, bielefeld2026signalpeakpowerconstraint}.
We denote the set of EDs as $\mathcal{K}=\{1,\ldots,K\}$.
The EDs share an uplink channel and simultaneously transmit over it.
Each ED is subject to the peak-power constraint.
As in previous studies~\cite{yang2022over,zhong2025over,azimi2024over,bielefeld2026signalpeakpowerconstraint,9272666}, we assume that the downlink is a reliable digital link, so that the model broadcast by the PS is received without error.
A communication round thus consists of i) local computation at the EDs, ii) simultaneous uplink AirComp aggregation, and iii) a downlink broadcast of the received aggregate. 
In the remainder of this section, we introduce learning and channel modeling of AirComp-FL systems.

\paragraph*{Learning modeling} 

In AirComp-FL systems, the EDs cooperatively train a neural network model, parameterized by $d$ trainable parameters denoted by $\mathbf{w}\in\mathbb{R}^d$. 
For training, each ED $k\in\mathcal{K}$ has a local dataset $\mathcal{D}_k$, where the datasets $\mathcal{D}_1,\ldots,\mathcal{D}_K$ do not overlap, i.e.,  $\mathcal{D}_i\cap\mathcal{D}_j=\emptyset$ for $i\ne j$. 
The local loss function of ED $k$ is written as 
\begin{equation}
    F_k(\mathbf{w}) = \frac{1}{|\mathcal{D}_k|}\sum_{\xi\in\mathcal{D}_k}\ell(\mathbf{w};\xi),
\end{equation}
where $\ell$ is the per-sample loss, e.g., cross-entropy.
The EDs cooperate with the PS to minimize the average of their local loss functions:
\begin{equation}
    \min_{\mathbf{w}\in\mathbb{R}^d}~F(\mathbf{w}) = \frac{1}{K}\sum_{k\in\mathcal{K}}F_k(\mathbf{w}),
\end{equation}
where we assume $|\mathcal{D}_i|=|\mathcal{D}_j|$ for all $i,j\in\mathcal{K}$ for notational brevity.
To solve the above problem efficiently and privately, EDs exchange local model update vectors instead of raw local data.
The PS then aggregates the updates and broadcasts the result to the EDs. 
Since the distribution of each local dataset $\mathcal{D}_i$ depends on where its data are generated, the local datasets are statistically heterogeneous in general.

\paragraph*{Channel modeling}

For uplink parameter transmission, each ED communicates with the PS by single-carrier AirComp with symbol-level synchronization, as in~\cite{yang2020federated,bielefeld2026signalpeakpowerconstraint, 8364613,zhong2025over,yang2022over,azimi2024over,oh2024communication,jang2024fed}.
For brevity, we consider a real-valued baseband single-input single-output (SISO) scenario, i.e., the EDs and PS are equipped with a single antenna, and we adopt a block-fading model in which the gain $h_{k,t}\in\mathbb{R}$ of ED $k$ is constant over the $m$ channel uses of round $t$. 
Over such a block, ED $k$ transmits $\mathbf{x}_{k,t}\in\mathbb{R}^{m}$ and the PS receives 
\begin{equation}\label{eq:channel_1}
    \mathbf{y}_t = \sum_{k\in\mathcal{K}}h_{k,t}\mathbf{x}_{k,t} + \bn_t,
\end{equation}
where $\mathbf{n}_t\sim\mathcal{N}(0,N_0\bI_m)$, and $N_0$ denotes the additive white Gaussian noise power at the PS.
Denoting the encoded block as $\mathbf{s}_{k,t}\in\mathbb{R}^m$, we follow the uniform-forcing principle of~\cite{yang2020federated,jang2024fed,8364613,9272666,bielefeld2026signalpeakpowerconstraint}, which gives a simplified channel model $\mathbf{x}_{k,t}=(c_t/h_{k,t})\mathbf{s}_{k,t}$ by channel inversion with a positive common scale $c_t>0$, \textit{regardless of the number of antennas}.\footnote{By solving the uniform-forcing beamforming problem as in~\cite{8364613,9272666,jang2024fed}, the multi-antenna cases are also represented by a Gaussian multiple-access channel.}
Then, the channel model in \eqref{eq:channel_1} can be reduced to
\begin{equation}\label{eq:channel_2}
    \mathbf{y}_t = c_t\sum_{k\in\mathcal{K}} \mathbf{s}_{k,t}+\mathbf{n}_t,
\end{equation}
i.e., a Gaussian multiple-access channel returning the unweighted sum of the encoded blocks, which is the property that AirComp exploits.
We note that, after \eqref{eq:channel_2}, the fading gains enter the remainder of this paper through the single scalar $c_t$, which is determined by the channel condition.

\paragraph*{Peak-power constraint and PAPR}

A practical PA is linear only up to a saturation level, beyond which its output compresses and the spectrum regrows out of band~\cite{cripps2006rf,rapp1991effects}.
Hence, the transmit block of ED $k$ at round $t$ must satisfy the \textit{peak power constraint}
\begin{equation}\label{eq:peak}
    \max_{i\in[m]} |[\mathbf{x}_{k,t}]_i|^2 = \left\Vert\mathbf{x}_{k,t}\right\Vert_\infty^2 \le P_\text{pk},
\end{equation}
where $P_\text{pk}$ denotes the maximum peak transmission power.
Any block can be scaled down to satisfy \eqref{eq:peak}; what the scaling costs is average transmitted power, by a factor given by the PAPR:
\begin{equation}\label{eq:papr}
    \PAPR_{k,t}= \frac{\left\Vert\mathbf{x}_{k,t}\right\Vert_\infty^2}{\tfrac{1}{m}\left\Vert\mathbf{x}_{k,t}\right\Vert_2^2}\in[1,m].
\end{equation}
Combining \eqref{eq:peak} and \eqref{eq:papr}, the average power a block delivers is at most $P_\text{pk}/\PAPR_{k,t}$.
Besides reducing the deliverable power, the back-off lowers the drain efficiency of every linear amplifier class~\cite{cripps2006rf,rapp1991effects}.
Hence, provisioning a larger back-off degrades the efficiency, and every chain is committed at design time to a finite back-off, as in~\cite{3gpp2023nr}. 
The PAPR is therefore not a constraint but the \textit{price} of the peak-power constraint, since it bounds the average power delivered.
Since \eqref{eq:papr} is invariant to scaling, it depends only on the \textit{shape} of the encoded signal; hence, we use a clipping operator to adjust its shape.


\section{Proposed Method: \AirGCCD}

In this section, we present \AirGCCD, which consists of i) Gaussian-circulant sketch/desketch and ii) clipping and debiasing. Fig.~\ref{fig:system_model} illustrates the end-to-end workflow from local update encoding at the EDs to over-the-air aggregation and local desketching after the PS broadcast. In AirComp-FL, the transmitted symbols are the entries of the model updates, which generally depend on the training data and the current model weights, and hence their distribution is analytically intractable.
To circumvent this, we propose an encoder-decoder design that guarantees a Gaussian distribution for the encoded updates, irrespective of the underlying data and model state, thereby rendering the subsequent analysis tractable.
We also show that Bussgang decomposition of the clipped blocks yields an unbiased aggregation despite clipping distortion, while the proposed design satisfies the peak power constraint~\eqref{eq:peak} and preserves the superposition property.

As in existing studies~\cite{yang2020federated,pmlr-v54-mcmahan17a,jang2024fed,qin2023federated}, the model parameters are initialized randomly and synchronized across EDs.
The EDs and PS cooperate for $T$ rounds, each of which consists of $I$ local updates.
In round $t$, the local parameter of ED $k$ in the local update step $i$ is denoted as $\mathbf{w}_{k,t,i}$.
The global weights at the $t$-th round are denoted by $\mathbf{w}_{\text{G},t}$. 
The initialized parameters are thus $\mathbf{w}_{k,0,0}=\mathbf{w}_{\text{G},0}, \forall k\in\mathcal{K}$.

\subsection{Local Update}

As per FedAvg~\cite{pmlr-v54-mcmahan17a}, at round $t$ every ED starts from the common model $\mathbf{w}_{\text{G},t}$, which is identical across EDs owing to the decoding in \cref{sec:decoding} and executes $I$ stochastic gradient descent (SGD) steps with a mini-batch $\mathcal{B}_{k,t,i}\subset\mathcal{D}_k$ of size $B$, i.e., starting from $\mathbf{w}_{k,t,0}$, the local parameter weight is updated by 
\begin{equation}\label{eq:local_sgd}
    \mathbf{w}_{k,t,i+1} = \mathbf{w}_{k,t,i} - \eta\mathbf{g}_{k,t,i}, \,\,i=0,\ldots,I-1,
\end{equation}
where $\mathbf{g}_{k,t,i}=\tfrac{1}{B}\sum_{\xi\in\mathcal{B}_{k,t,i}}\nabla_\mathbf{w}\ell(\mathbf{w}_{k,t,i},\xi)$.
After $I$ local SGD steps, ED $k$ obtains its local model update vector as
\begin{equation}
    \Delta \mathbf{w}_{k,t} = \mathbf{w}_{k,t,I}-\mathbf{w}_{k,t,0}=-\eta \sum_{i=0}^{I-1} \mathbf{g}_{k,t,i}\in\mathbb{R}^d.
\end{equation}
In AirComp-FL, all participating EDs deliver the local updates $\Delta\mathbf{w}_{k,t}$ simultaneously over the uplink channel \eqref{eq:channel_2}. 
Then, the PS broadcasts the received vector $\by_t$ to the EDs via reliable digital communication links. 

\subsection{Encoding: Gaussian-Circulant Precoding}

In canonical AirComp-FL, directly transmitting $\Delta\mathbf{w}_{k,t}$ i) occupies $d$ channel uses and ii) incurs a PAPR of up to $d$, both of which stem from transmitting the raw update itself.
To address these limitations, we adopt vector \textit{sketching}~\cite{sarlos2006improved}, a \textit{random} linear transform that compresses the update while reducing its PAPR.
Several sketching schemes have been proposed for AirComp-FL~\cite{jang2024fed, 10472332}, including the Gaussian sketch, the SRHT-based sketch, and the count sketch.
The Gaussian sketch is analytically tractable, as its output is exactly Gaussian, but requires $\mathcal{O}(md)$ operations, where $m$ denotes the sketch dimension.
The SRHT-based and count sketches, in contrast, yield biased reconstructed updates when combined with clipping, but are computationally cheaper, e.g., $\mathcal{O}(d\log d)$ operations for the SRHT-based sketch.

\paragraph*{Partial Gaussian circulant sketch}

In this paper, we introduce a \textit{partial Gaussian circulant sketch}, whose output is exactly Gaussian while incurring only $\mathcal{O}(d\log d)$ complexity.
Since the proposed method relies on the radix-2 fast Fourier transform (FFT) to reduce the complexity, each ED zero-pads the update vector $\Delta\bw_{k,t}$ to a power-of-two dimension.
Denoting the zero-padded dimension as $\bar{d}=2^{\lceil\log_2d\rceil}$, the zero-padding multiplies the update vector by the matrix $\bJ\coloneqq[\bI_d | \bO_{d,\bar{d}-d}]^\mathrm{\top}\in\{0,1\}^{\bar{d}\times d}$, where $\bO_{d,\bar{d}-d}$ denotes the $d\times (\bar{d}-d)$ all-zero matrix.
Its transpose $\bJ^\top$ truncates a $\bar{d}$-length vector to its first $d$ entries and $\norm{\bJ\bx}=\norm{\bx}$.
Let $\bd\in\{-1,1\}^{\bar{d}}$ be a sign vector with i.i.d. Rademacher entries, and let $\bD=\mathtt{diag}(\bd)\in\{-1,0,1\}^{\bar{d}\times\bar{d}}$ be the corresponding diagonal sign matrix.
The Gaussian circulant matrix $\bC$ is generated by a zero-mean unit-variance Gaussian random vector $\bc\sim\mathcal{N}(0,\bI_{\bar{d}})$, whose cyclic shifts are defined by $[\bc^{(i)}]_j=[\bc]_{(i-j) \bmod\bar d +1}$. 
Then, the Gaussian circulant matrix $\bC$ is defined by 
\begin{equation}\label{eq:circ}
    [\bC]_{i,j} = [\bc^{(i)}]_j.
\end{equation}
In addition to the aforementioned matrices, we consider a subsampling matrix $\bP_\Omega\in\{0,1\}^{m\times\bar{d}}$, where $\Omega\subset[\bar{d}]$ with $|\Omega|=m$ is drawn uniformly \textit{without replacement}.
The sketch matrix is then defined as 
\begin{equation}\label{eq:sketchmat}
    \bPhi_t=\frac{1}{\sqrt{m}}\bP_\Omega\bC \bD \bJ\in\mathbb{R}^{m\times d},
\end{equation}
where the random matrices $\bP_\Omega$, $\bC$, and $\bD$ are generated by a common random seed $\psi_t$.
Then, the resulting sketch output before clipping and rescaling is
\begin{equation}\label{eq:sketch_output}
    \bar{\bs}_{k,t}=\bPhi_t\Delta\bw_{k,t}\in\mathbb{R}^{m}.
\end{equation}

\paragraph*{Fast implementation via the FFT}

In \eqref{eq:sketch_output}, since $\bP_\Omega$ is a sampling matrix, $\bD$ is a diagonal sign matrix, and $\bJ$ is a rectangular zero-padding matrix, multiplying them with a vector requires $\mathcal{O}(\bar{d})$ operations.
However, the Gaussian circulant matrix $\bC$ is a dense matrix; hence, directly evaluating the $m$ sampled entries in \eqref{eq:sketch_output} requires $\mathcal{O}(md)$ operations.
In this paper, we reduce the computational complexity of the matrix-vector product of $\bC$ by diagonalizing it with the unitary DFT matrix $\bF$ of order $\bar{d}$, i.e., 
\begin{equation}\label{eq:diag}
    \bC=\bF^{-1}\mathtt{diag}\big(\widehat{\bc}\big)\bF,
\end{equation}
where $[\bF]_{lp} = e^{-j2\pi (l-1)(p-1)/\bar{d}}/\sqrt{\bar{d}}$.
Because $\mathtt{FFT}(\bx)=\sqrt{\bar{d}} \bF \bx$ and $\mathtt{IFFT}(\bx)=\sqrt{\tfrac{1}{\bar{d}}}\bF^{-1}\bx$, we simplify \eqref{eq:sketch_output} by replacing $\bC$ with its diagonalized form as follows:
\begin{align}\label{eq:fft}
    \bPhi_t \Delta\bw_{k,t} & = \frac{1}{\sqrt{m}}\bP_\Omega \bF^{-1} \mathtt{diag}(\widehat{\bc})\bF \bD\bJ\Delta\bw_{k,t} \\ 
    \nonumber & = \frac{1}{\sqrt{m}} \,[\mathtt{IFFT}(\widehat{\bc}\odot\mathtt{FFT}(\bd\odot\bJ\Delta\bw_{k,t}))]_{\Omega},
\end{align}
where $\odot$ denotes the Hadamard product.
Since $\bar{d}$ is a power of two, $\bF$ admits the radix-2 factorization. 
Denoting the unitary DFT matrix of order $\bar{d}/2$ by $\bF_{\bar{d}/2}$, the matrix-vector product $\bF\bx$ can be decomposed as 
\begin{equation}\label{eq:radix2}
    \bF\bx = \frac{1}{\sqrt{2}}\begin{bmatrix}
        \bI_{\bar{d}/2} & \bA_{\bar{d}/2}\\
        \bI_{\bar{d}/2} & -\bA_{\bar{d}/2}
    \end{bmatrix}
    \begin{bmatrix}
        \bF_{\bar{d}/2} \bx_{o}\\
        \bF_{\bar{d}/2} \bx_{e} \\
    \end{bmatrix},
\end{equation}
where $\bx_o=[x_1,x_3,\cdots,x_{\bar{d}-1}]^\top$, $\bx_e=[x_2,x_4,\cdots,x_{\bar{d}}]^\top$, and $\bA_{\bar d/2}=\mathtt{diag}(1,e^{-\mathrm{j}2\pi/\bar d},\dots,e^{-\mathrm{j}2\pi(\bar{d}/2-1)/\bar d})$.
In \eqref{eq:radix2}, the butterfly combination of the two half-size DFT outputs requires $\mathcal{O}(\bar{d})$ operations. 
Thus, by recursion of  \eqref{eq:radix2}, the total computational complexity of $\bF\bx$ is $\mathcal{O}(\bar d\log\bar d)=\mathcal{O}(d\log d)$, since $\bar d<2d$.
In Alg. \ref{alg:radix2}, we state the recursion in  $\FFT(\bx)=\sqrt{\bar d}\,\bF\bx$ used by \eqref{eq:fft}, with the inverse obtained by conjugation, $\IFFT(\bx)=\big(\FFT(\bx^{*})\big)^{*}/\bar d$, where $^*$ denotes conjugation.

\begin{algorithm}[!t]
\caption{Radix-$2$ FFT and IFFT.}
\label{alg:radix2}
\SetKwProg{Fn}{Function}{:}{}

\Fn{$\FFT(\bx\in\mathbb{C}^{n})$}{
    \lIf{$n=1$}{\KwRet{$\bx$}}
    $\widehat\bx_{\mathrm{e}}\leftarrow\FFT([x_0,x_2,\dots,x_{n-2}])$;\quad
    $\widehat\bx_{\mathrm{o}}\leftarrow\FFT([x_1,x_3,\dots,x_{n-1}])$\;
    \For{$l=0,\dots,n/2-1$}{
        $t\leftarrow e^{-\mathrm{j}2\pi l/n}\,[\widehat\bx_{\mathrm{o}}]_l$ \tcp*{$\bA$ of \eqref{eq:radix2}}
        $[\widehat\bx]_l\leftarrow[\widehat\bx_{\mathrm{e}}]_l+t$;\qquad
        $[\widehat\bx]_{l+n/2}\leftarrow[\widehat\bx_{\mathrm{e}}]_l-t$ 
    }
    \KwRet{$\widehat\bx$}
}
\vspace{2pt}

\Fn{$\IFFT(\bx\in\mathbb{C}^{n})$}{
    \KwRet{$\big(\FFT(\bx^{*})\big)^{*}/\,n$} \tcp*{$(\cdot)^{*}$: entrywise conjugate}
}
\end{algorithm}

\paragraph*{Distribution of $\widehat{\bc}$}

In \eqref{eq:fft}, the diagonalizable Gaussian circulant matrix $\bC$ enables the matrix-vector product $\bPhi_t\Delta\bw_{k,t}$ within $\mathcal{O}(d\log d)$ operations, without generating the matrix $\bPhi_t$, which we call \textit{fast sketch}.
However, the fast sketch requires $\widehat{\bc}$ instead of $\bc$; thus, we derive the distribution of $\widehat{\bc}$ in \eqref{eq:diag}.
To this end, we have $\bF\bC = \mathtt{diag}(\widehat{\bc})\bF$ by rearranging \eqref{eq:diag}. Comparing the first column of both sides of the above equation, we have
\begin{equation}\label{eq:eigen_value}
    [\widehat{\bc}]_l=\sum_{p=1}^{\bar d} [\bc]_p\,e^{-j2\pi (l-1)(p-1)/\bar d},
\end{equation}
which can be rewritten as $\widehat{\bc}=\sqrt{\bar{d}}\bF\bc$.
The independent entries in the first half-spectrum of $\widehat{\bc}$ have the distributions 
\begin{equation}\label{eq:rhat}
    [\widehat{\bc}]_l \sim \begin{cases}
        \mathcal{N}(0,\bar{d})& \text{i.i.d.},~l=1,\bar{d}/2+1\\ 
        \mathcal{CN}(0,\bar{d})&\text{i.i.d.},~\text{o.w.},
    \end{cases}
\end{equation}
where $\mathcal{CN}$ denotes the complex Gaussian distribution. The remaining entries satisfy $[\widehat{\bc}]_{\bar d-l+2}=[\widehat{\bc}]_l^*$ for $l=2,\ldots,\bar d/2$, so that $\widehat{\bc}\in\mathbb{C}^{\bar d}$ has conjugate symmetry and a real IFFT.
Using the distribution of $\widehat{\bc}$ in \eqref{eq:rhat}, the sketch output \eqref{eq:sketch_output} can be obtained without an explicit matrix-vector product, as in Alg.~\ref{alg:fft}.
We note that the desketch procedure is addressed in \cref{sec:decoding}.

\begin{algorithm}[!t]
\caption{Circulant Sketch and Desketch of \AirGCCD.}
\label{alg:fft}
\SetKwFunction{FSketch}{Sketch}
\SetKwFunction{FDesketch}{Desketch}
\SetKwProg{Fn}{Function}{:}{}

\KwIn{Model size $d$, block length $m$, shared seed $\psi_t$.\!\!}
\textbf{Initialization (shared by all EDs and the PS via seed $\psi_t$):}\;
$\bar d\leftarrow2^{\lceil\log_2 d\rceil}$\;
Generate $\bd\in\{\pm1\}^{\bar d}$ and $\Omega\subset[\bar d]$, $|\Omega|=m$, from $\psi_t$.\;
Generate $\widehat{\bc}\in\mathbb{C}^{\bar d}$ from $\psi_t$ by \eqref{eq:rhat} and conjugate symmetry \tcp*{$\mathcal{O}(\bar d)$}
\vspace{2pt}

\Fn{\FSketch{$\Delta\bw_{k,t}$}}{
    \tcp{Executed at each ED $k$}
    $\bz\leftarrow\bd\odot\big[\Delta\bw_{k,t};\mathbf{0}_{\bar d-d}\big]$ \tcp*{pad and sign-flip, $\mathcal{O}(\bar d)$}
    $\bz\leftarrow\mathrm{IFFT}\big(\widehat{\bc}\odot\mathrm{FFT}(\bz)\big)$ \tcp*{Alg. \ref{alg:radix2}, $\mathcal{O}(d\log d)$}
    \KwRet{$[\bz]_\Omega/\sqrt m\in\mathbb{R}^{m}$ \tcp*{$\bar{\bs}_{k,t}=\bPhi_t\Delta\bw_{k,t}$}}
}
\vspace{2pt}

\Fn{\FDesketch{$\by_t$}}{
    \tcp{Executed at every ED after the broadcast}
    $\bz\leftarrow\mathbf{0}_{\bar d}$;\quad $[\bz]_\Omega\leftarrow\by_t/\sqrt m$ \tcp*{scatter}
    $\bz\leftarrow\mathrm{IFFT}\big(\widehat{\bc}^{*}\odot\mathrm{FFT}(\bz)\big)$ \tcp*{$\bC^\top\bz$}
    \KwRet{first $d$ entries of $\bd\odot\bz$ \tcp*{$\bPhi_t^\top\by_t$, truncation by $\bJ^\top$}}
}
\end{algorithm}

\subsection{Clipping and Bussgang Decomposition}

In \eqref{eq:sketch_output}, we have the sketch output $\bar{\bs}_{k,t}$, which is obtained with $\mathcal{O}(d\log d)$ operations. 
As we mentioned before, we aim to find the optimal clipping threshold for the sketch output $\bar{\bs}_{k,t}$.
Because the PAPR in \eqref{eq:papr} is invariant to the power scaling, we adopt the simplest mechanism: each ED \textit{clips} its precoded block \eqref{eq:sketch_output} before transmission as $\clip_{\tau_{k,t}}(\bar\bs_{k,t})$,
where $\tau_{k,t}$ denotes the clipping level of ED $k$ at communication round $t$.

\paragraph*{Bussgang decomposition}

Although clipping reduces the PAPR, it distorts the transmit block, and this distortion may bias the aggregation after superposition.
Thus, we first split $\clip_{\tau_{k,t}}(\bar\bs_{k,t})$ into its best linear approximation and a residual term to analyze the gradient bias.
Following Bussgang~\cite{bussgang1952crosscorrelation}, the clipping output can be decomposed as 
\begin{equation}\label{eq:split}
    \clip_{\tau_{k,t}}(\bar\bs_{k,t}) = C_{\text{Buss}}\bar\bs_{k,t} + f_{k,t}(\bar\bs_{k,t}),
\end{equation}
where $C_{\text{Buss}}$ is the Bussgang gain defined by  
\begin{equation}\label{eq:alphaproj}
    C_{\text{Buss}} = \frac{\mathbb{E}\left[[\bar\bs_{k,t}]_i\cdot \clip_{\tau_{k,t}}([\bar\bs_{k,t}]_i)\right]}{\mathbb{E}\left[[\bar\bs_{k,t}]_i^2\right]}
\end{equation}
and $f_{k,t}(\bar\bs_{k,t})$ is the residual term.
To evaluate the Bussgang gain, we first derive the distribution of each entry of $\bar{\bs}_{k,t}$ \eqref{eq:sketch_output}.

\begin{lemma}[Gaussian coordinates]
\label{lem:be}
    For any $\Delta\bw_{k,t}\neq\mathbf{0}$, we let $[\bar\bs_{k,t}]_i$ be the corresponding sketch coordinate in \eqref{eq:sketch_output}. Conditioned on $\bD$, for every $i\in[m]$, we have 
    \begin{equation}
        [\bar\bs_{k,t}]_i\sim\mathcal{N}(0,\sigma_s^2),
    \end{equation}
    where $\sigma_s^2=\Vert\Delta\bw_{k,t}\Vert^2/m$.
    The proof appears in Appendix~\ref{app:be}.
\end{lemma}

In \cref{lem:be}, we show that the output of the proposed sketch \eqref{eq:sketch_output} is Gaussian.
Using this Gaussianity, \cref{lem:bussgang} evaluates the numerator of \eqref{eq:alphaproj}, together with the second moments used in the bias analysis.
\begin{lemma}[Statistics of a Gaussian block]
\label{lem:bussgang}
    Let $\varsigma\sim\mathcal{N}(0,\sigma_s^2)$ be a coordinate of a Gaussian block, and let $\gamma=\tau/\sigma_s$ and $f(x) = \clip_\tau(x) - \alpha_\gamma x$.
    Denoting $\alpha_\gamma=\int_{-\gamma}^\gamma\tfrac{1}{\sqrt{2\pi}}e^{-x^2/2}dx=\mathrm{erf}(\gamma/\sqrt{2})$, $\omega(\gamma)=\alpha_\gamma-2\gamma\phi(\gamma)+2\gamma^2Q(\gamma)$, and $A(\gamma)=\omega(\gamma)-\alpha_\gamma^2$, we have
    \begin{equation}\ \label{eq:zclip}
    \begin{cases}
        \mathbb{E}\big[\varsigma\clip_\tau(\varsigma)\big]&=\alpha_\gamma\sigma_s^2, \\
        \mathbb{E}\big[\clip_\tau(\varsigma)^2\big]&=\omega(\gamma)\,\sigma_s^2, \\
        \mathbb{E}\big[f(\varsigma)^2\big]&=A(\gamma)\,\sigma_s^2.
    \end{cases}
    \end{equation}
    The proof appears in Appendix~\ref{app:gauss}.
\end{lemma}

By using \eqref{eq:zclip} in \cref{lem:bussgang} and defining a common clipping ratio $\gamma=\tau_{k,t}/\sigma_s$, i.e., $\tau_{k,t}=\gamma\sigma_s$, we have
\begin{equation}\label{eq:alpha_buss}
    C_{\text{Buss}} = \frac{\alpha_{\gamma}\sigma_s^2}{\sigma_s^2} = \alpha_{\gamma}.
\end{equation}
Then, by dividing \eqref{eq:split} by $\alpha_{\gamma}$, the transmitted block $\bs_{k,t}$ in \eqref{eq:channel_2} is given by
\begin{equation}\label{eq:x_kt}
    \bs_{k,t} = \frac{\clip_{\tau_{k,t}}(\bar\bs_{k,t})}{\alpha_\gamma} =\bPhi_t\Delta\bw_{k,t} + \frac{f_{k,t}(\bar\bs_{k,t})}{\alpha_{\gamma}}.
\end{equation}
The clipping design leaves two questions, which \cref{sec:theoretical} addresses: i) at which level to clip, and ii) what statistical distortion the clipping incurs.
Before the theoretical analysis, we introduce the aggregation and decoding procedures.

\subsection{Aggregate Broadcast and Decoding at the EDs}
\label{sec:decoding}
Substituting \eqref{eq:x_kt} into \eqref{eq:channel_2}, the received block at the PS is denoted as
\begin{equation}\label{eq:superpos}
    \by_t
    = c_t\underbrace{\,\bPhi_t\Big(\sum_{k\in\mathcal{K}}\Delta\bw_{k,t}\Big)}_{\text{Sketch}}
    + c_t\underbrace{\sum_{k\in\mathcal{K}}\frac{f_{k,t}(\bar\bs_{k,t})}{\alpha_{\gamma}}}_{\text{Residual}}
    + \bn_t,
\end{equation}
which consists of the \textit{sum of} sketches, \textit{sum of} residual terms, and channel noise.
After receiving $\by_t$, the PS rescales it to $\by_t/(c_tK)$ and broadcasts it to the EDs over the reliable digital link. 

\paragraph*{Desketch}

After receiving $\by_t/(c_tK)$ from the PS, each ED locally applies the desketch matrix $\bPhi_t^\top$, which it generates from the same seed $\psi_t$.
Then, the desketch output $\bPhi_t^\top \by_{t}/(c_tK)$ is given by 
\begin{align}\label{eq:decoder}
    \Delta\widehat{\bw}_{G,t}
    &=\frac{1}{c_t K}\bPhi_t^\top\by_t \nonumber\\
    &=\frac{1}{K}\sum_{k\in\mathcal{K}}\Delta\bw_{k,t} + \frac{1}{K}\sum_{k\in\mathcal{K}}(\bPhi_t^\top\bPhi_t-\mathbf{I}_d)\Delta\bw_{k,t} \\ 
    & +\frac{1}{K}\sum_{k\in\mathcal{K}}\frac{\bPhi_t^\top f_{k,t}(\bar\bs_{k,t})}{\alpha_{\gamma}} 
    +\frac{1}{c_tK}\bPhi_t^\top\bn_t.\nonumber
\end{align}
The decoded update in \eqref{eq:decoder} is identical across all EDs, keeping their local models synchronized without model broadcast; its deviation from $\tfrac{1}{K}\sum_k\Delta\bw_{k,t}$ consists of sketching, clipping, and channel distortions.
At the end of round $t$, each ED updates its model locally by
\begin{equation}
    \bw_{G,t+1} = \bw_{G,t} + \Delta\widehat{\bw}_{G,t}.
\end{equation}

\subsection{Power Control under the Peak Power Constraint}

In \eqref{eq:peak}, the peak power of $\bx_{k,t}$ is limited by $P_\text{pk}$, and the uniform-forcing gives us $\bx_{k,t}=(c_t/h_{k,t})\bs_{k,t}$. 
Hence, the peak power constraint can be rewritten as 
\begin{equation}\label{eq:peak_rewritten}
    \frac{c_t^2 \min\{\Vert\bar\bs_{k,t}\Vert_\infty^2,\tau_{k,t}^2\}}{h_{k,t}^2\alpha_{\gamma}^2} \le P_{\text{pk}},~~ \forall k\in\mathcal{K}.
\end{equation}    
Thus, the largest common scale $c_t$ satisfying $\max_{k}\Vert\bx_{k,t}\Vert_\infty^2 = P_{\text{pk}}$ is obtained by
\begin{equation}\label{eq:cstar}
    c_t^\star=  \sqrt{P_{\text{pk}} \cdot\min_{k\in\mathcal{K}} \frac{h_{k,t}^2\alpha_{\gamma}^2}{\min\{\Vert\bar\bs_{k,t}\Vert_\infty^2,\tau_{k,t}^2\}}}.
\end{equation}
Finally, Alg.~\ref{alg:aircirc} gives the implementation details of \AirGCCD.

\begin{algorithm}[!t]
\caption{\AirGCCD Algorithm (round $t$).}
\label{alg:aircirc}
\KwIn{Global model $\bw_{G,t}$, clipping ratio $\gamma$ (\cref{cor:opt}), peak budget $P_{\mathrm{pk}}$, learning rate $\eta$, local steps $I$.}
\KwOut{Updated global model $\bw_{G,t+1}$.}
\underline{\textsc{On PS (downlink):}} broadcast a fresh seed $\psi_t$ (and $\bw_{G,0}$ once at $t=0$).\;
\underline{\textsc{On EDs (local training and encoding):}}\;
\ParallelFor(){\textbf{ED $k\in\mathcal{K}$}}{
    Run $I$ local SGD steps as in \eqref{eq:local_sgd} and obtain $\Delta\bw_{k,t}$\;
    $\bar\bs_{k,t}\leftarrow\bPhi_t\Delta\bw_{k,t}$ \tcp*{Alg. \ref{alg:fft}}
    $\tau_{k,t}\leftarrow\gamma\,\Vert\Delta\bw_{k,t}\Vert/\sqrt m$ \;
    Send the quantized scalar $\min\{\Vert\bar\bs_{k,t}\Vert_\infty^2,\tau_{k,t}^2\}$ and CSI to the PS (digital).\;
}
\underline{\textsc{On PS (power control):}} $c_t\leftarrow c_t^\star$ as in \eqref{eq:cstar}; feed back $c_t$ (digital).\;
\underline{\textsc{On EDs (transmission):}}\;
\ParallelFor(){\textbf{ED $k\in\mathcal{K}$}}{
    $\bs_{k,t}\leftarrow\clip_{\tau_{k,t}}(\bar\bs_{k,t})/\alpha_\gamma$ \;
    Transmit $\bx_{k,t}=\frac{c_t}{h_{k,t}}\bs_{k,t}$ over $m$ channel uses.\;
}
\underline{\textsc{On PS (aggregation and broadcast):}}\;
Receive $\by_t$ as in \eqref{eq:superpos}; broadcast $\by_t/(c_tK)$ (digital, $m$ symbols).\;
\underline{\textsc{On EDs (decoding):}}\;
\ParallelFor(){\textbf{ED $k\in\mathcal{K}$}}{
    $\Delta\widehat{\bw}_{G,t}\leftarrow\bPhi_t^\top\by_t/(c_tK)$ \tcp*{Alg. \ref{alg:fft}}
    $\bw_{G,t+1}\leftarrow\bw_{G,t}+\Delta\widehat{\bw}_{G,t}$.\;
}
\end{algorithm}


\section{Theoretical Analysis}
\label{sec:theoretical}

In this section, we theoretically analyze i) the bias and variance of the decoded update~\eqref{eq:decoder}, and ii) the convergence of \AirGCCD in Alg. \ref{alg:aircirc}. 
To this end, we reuse \cref{lem:bussgang,lem:be} and first analyze the moments of the sketch-desketch procedure.
\begin{lemma}[Sketch moments]
\label{lem:sketch}
    For any fixed $\bu\in\mathbb{R}^{d}$ and $\bs=\bPhi_t\bu$, the following statements hold.
    \begin{enumerate}[label=(\roman*), leftmargin=2.5em]
        \item $\mathbb{E}[\bPhi_t^\top\bPhi_t]=\bI_{d}$.
        \item $\mathbb{E}\Vert(\bPhi_t^\top\bPhi_t-\bI_{d})\bu\Vert^2\le\left(\frac{d+1}{m}+\frac{1}{\bar{d}-1}\right)\Vert\bu\Vert^2$.
        \item $\mathbb{E}\Vert\bPhi_t^\top\bn\Vert_2^2=dN_0$ for $\bn\sim\mathcal{N}(\boldsymbol{0}, N_0\bI_m)$. 
    \end{enumerate}
    The proof is shown in Appendix~\ref{app:sketch}.
\end{lemma}

Parts (i) and (ii) of \cref{lem:sketch} give the mean and variance of the sketch-desketch procedure without clipping. 
Part (iii) shows that the desketched channel noise has total variance $dN_0$, exactly the same as in uncompressed AirComp-FL. 

\subsection{Aggregation Bias}
\label{sec:bias}

Taking the expectation of \eqref{eq:decoder} over ($\bD$, $\bC$, $\bP_\Omega$, $\bn_t$), the bias of the aggregated update is given by
\begin{equation}\label{eq:meanest}
    \mathbb{E}\big[\Delta\widehat{\bw}_{G,t}\big] - \frac{1}{K}\sum_{k\in\mathcal{K}} \Delta\bw_{k,t}
    = \frac{1}{K}\sum_{k\in\mathcal{K}}\bb_{k,t}, 
\end{equation}
where
\begin{equation}\label{eq:bb_kt}
    \bb_{k,t}=\mathbb{E}\left[\bPhi_t^\top f_{k,t}(\bar\bs_{k,t})/\alpha_{\gamma}\right].
\end{equation}
We first characterize the mean aggregation error in \eqref{eq:meanest}. 
Unlike zero-mean perturbations, systematic bias can persist during gradient descent~\cite{ajalloeian2020convergence}, and $\bb_{k,t}$ in \eqref{eq:bb_kt} represents the bias induced by clipping.

\begin{theorem}[Unbiased aggregation of \AirGCCD]
\label{thm:bias}
    Let each ED transmit \eqref{eq:x_kt} at the clipping level of $\tau_{k,t}=\gamma\sigma_s$.
    Then, the aggregated update in \eqref{eq:decoder} is unbiased, i.e., 
    \begin{equation}\label{eq:biasfree}
        \bb_{k,t}=\mathbf{0}.
    \end{equation}
    The proof can be found in Appendix~\ref{app:bias}.
\end{theorem}

\cref{thm:bias} shows that the clipping bias $\bb_{k,t}$ is zero, so the decoded update in \eqref{eq:decoder} is unbiased and the clipping distortion enters only through the variance, which we analyze next.

\subsection{Variance of the Recovered Update}
\label{sec:moments}

Subtracting $\frac{1}{K}\sum_{k\in\mathcal{K}}\Delta\bw_{k,t}$ from \eqref{eq:decoder} splits the error into three parts,
\begin{align}\label{eq:threeterms}
    \Delta\widehat{\bw}_{G,t}-\frac{1}{K}\sum_{k\in\mathcal{K}}\Delta\bw_{k,t}
    &=\underbrace{\tfrac{1}{K}(\bPhi_t^\top\bPhi_t-\bI_d)\textstyle\sum_{k\in\mathcal{K}}\Delta\bw_{k,t}}_{=\bee_{\mathrm{sk}}} \nonumber\\
    &\quad+\underbrace{\tfrac{1}{K\alpha_\gamma}\textstyle\sum_{k\in\mathcal{K}}\bPhi_t^\top f_{k,t}(\bPhi_t\Delta\bw_{k,t})}_{=\bee_{\mathrm{cl}}} \nonumber\\
    &\quad+\underbrace{\tfrac{1}{c_tK}\bPhi_t^\top\bn_t}_{=\bee_{\mathrm{ch}}},
\end{align}
where $\bee_{\mathrm{sk}}$, $\bee_{\mathrm{cl}}$, and $\bee_{\mathrm{ch}}$ denote the errors caused by sketching, clipping, and channel noise, respectively.

\begin{proposition}[Variance of the recovered update]
\label{prop:moments}
    Write $\Esk=\mathbb{E}[\Vert\bee_{\mathrm{sk}}\Vert^2]$, $\Ecl=\mathbb{E}[\Vert\bee_{\mathrm{cl}}\Vert^2]$, and $\Ech=\mathbb{E}[\Vert\bee_{\mathrm{ch}}\Vert^2]$, and let $c_t=c_t^\star$ as in \eqref{eq:cstar}.
    Then
    \begin{equation}\label{eq:moments}
        \mathbb{E}\big\Vert\Delta\widehat{\bw}_{G,t}-\textstyle\frac{\sum_{k}\Delta\bw_{k,t}}{K}\big\Vert^2\le\big(\sqrt{\Esk}+\sqrt{\Ecl}\big)^2+\Ech,
    \end{equation}
    since $\bee_{\mathrm{ch}}$ is uncorrelated with the other two terms, where
    \begin{equation}\label{eq:esk_cl_ch}
    \begin{cases}
        \Esk&\le\Delta_\mathrm{sk}\mathbb{E}\big[\big\Vert\textstyle\tfrac{1}{K}\sum_{k}\Delta\bw_{k,t}\big\Vert^2\big],\\
        \Ecl&\le\Delta_\mathrm{cl}\sum_{k}\frac{\mathbb{E}[\Vert\Delta\bw_{k,t}\Vert_2^2]}{K},\\
        \Ech&\le\Delta_\mathrm{ch}\max_k\frac{\mathbb{E}[\Vert\Delta\bw_{k,t}\Vert_2^{2}]}{K},
    \end{cases}
    \end{equation}
    where $A(\cdot)$ is defined in \cref{lem:bussgang},  
    \begin{equation}\label{eq:deltadef}
        \begin{cases}
            \Delta_{\mathrm{sk}}=\frac{d+1}{m}+\frac{1}{\bar d-1}, \\ 
            \Delta_{\mathrm{cl}}=\frac{d}{m}\frac{A(\gamma)}{\alpha_{\gamma}^{2}}\big(1+\varepsilon_{\mathrm{cl}}\big),\\
            \Delta_{\mathrm{ch}}=\frac{d\,\gamma^{2}N_0}{m\,\alpha_{\gamma}^{2}K P_{\mathrm{pk}}}\,\mathbb{E}\Big[\max_{k}h_{k,t}^{-2}\Big]
        \end{cases}
    \end{equation}
    and
    \begin{align}
        \varepsilon_{\mathrm{cl}}=\frac{1}{d\,A(\gamma)}\Big[&2(1-\alpha_{\gamma})\big(\alpha_{\gamma}-2\gamma\phi(\gamma)\big)\nonumber\\
        &+\big(4\alpha_{\gamma}(1-\alpha_{\gamma})+8\gamma\phi(\gamma)\big)\tfrac{m}{\bar d-1}\Big].
    \end{align} 
    The proof can be found in Appendix~\ref{app:mse}.
\end{proposition}

\cref{prop:moments} bounds the power of the three error terms of \eqref{eq:threeterms}:
\begin{itemize}
    \item The sketching loss (i.e., compression loss) $\Esk$ comes from \cref{lem:sketch} for the \textit{average} update, and the clipping ratio is chosen independently of this error. 
    \item The clipping loss $\Ecl$ carries the clipped Gaussian power $A(\gamma)$ in \cref{lem:bussgang}. This loss \textit{decays} super-exponentially in $\gamma$, since $A(\gamma)\approx 4\phi(\gamma)/\gamma^3$.
    \item In contrast, the channel noise $\Ech$ grows only quadratically in $\gamma$.
\end{itemize}
The asymmetry between the last two is what a design may exploit: the clipping ratio trades a super-exponentially small distortion against a quadratically growing noise, at a ratio the receive SNR fixes.

\subsection{Convergence Analysis}
\label{sec:convergence}

\cref{prop:moments} bounds the per-round error; we now translate this bound into a convergence rate.
We work under the following standard assumptions~\cite{ajalloeian2020convergence,jang2024fed}.

\begin{assumption}
\label{assumption:1}
    The local loss functions $F_k$ and the global loss $F$ satisfy the following.
    \begin{enumerate}[label=(A\arabic*), leftmargin=3em]
        \item $F$ is finite and lower-bounded, \ie, $F(\bw)\ge F^\star>-\infty$ for all $\bw\in\mathbb{R}^d$.
        \item For all $k\in\mathcal{K}$, $\bw$, and $\bw'$, $\Vert\nabla F_k(\bw')-\nabla F_k(\bw)\Vert\le\beta\Vert\bw'-\bw\Vert$.
        \item The mini-batch gradient is unbiased with bounded variance, \ie, $\mathbb{E}[\bg_{k,t,i}]=\nabla F_k(\bw_{k,t,i})$ and $\mathbb{E}\Vert\bg_{k,t,i}-\nabla F_k(\bw_{k,t,i})\Vert^2\le\xi^2$, and the mini-batches are drawn independently across EDs.
        \item $\Vert\nabla F_k(\bw)\Vert^2\le G_1^2$ for all $k$ and $\bw$, so that (A3) gives $\mathbb{E}\Vert\bg_{k,t,i}\Vert^2\le\xi^2+G_1^2\eqqcolon G_2^2$.
    \end{enumerate}
\end{assumption}

We allow \(I\ge1\) local steps per round and define
$ \bar{\bw}_{G,t,i}=\frac1K\sum_k\bw_{k,t,i}$
as the virtual average iterate with $\bar{\bw}_{G,t,0}=\bw_{G,t}$.

\begin{theorem}[Convergence]
\label{thm:conv}
    Under Assumption \ref{assumption:1}, if $\eta\ =\frac{1}{\sqrt{TI}}\le\ \tfrac{1}{\beta\big(1+2I\Delta_{\mathrm{sk}}\big)},$
    then \AirGCCD satisfies
    \begin{align}\label{eq:convrate}
        &\frac{1}{TI}\sum_{t=0}^{T-1}\sum_{i\in[I]}\mathbb{E}\big\Vert\nabla F(\bar\bw_{G,t,i-1})\big\Vert^{2} \\
        &\le\frac{1}{\sqrt{TI}}\Big[2\Delta_F+\underbrace{\frac{\beta\xi^{2}}{K}(1+2I\Delta_\mathrm{sk})}_{\text{mini-batch/sketch}}+\underbrace{\beta I\big(2\Delta_{\mathrm{cl}}+\Delta_{\mathrm{ch}}\big)G_2^{2}}_{\text{clipping/channel}}\Big] \nonumber\\
        &~~~+\underbrace{\frac{\beta^{2}(I-1)G_2^{2}}{3T}}_{\text{client drift}}.\nonumber
    \end{align}
    Here, $\Delta_F=F(\bw_{G,0})-F^\star$. The proof can be found in Appendix~\ref{app:conv}.
\end{theorem}

\cref{thm:conv} shows that the bound in \eqref{eq:convrate} decays at least at a rate of $1/\sqrt{T}$. 
More importantly, the learning rate $\eta$ must satisfy $\eta\le\frac{1}{\beta(1+2I(\frac{d+1}{m}+\frac{1}{\bar{d}-1}))}=\frac{1}{\beta(1+2I\Delta_{\textrm{sk}})}$, so the only distortion that tightens the \textbf{\textit{admissible learning rate} is the \textit{sketch distortion term}}: a larger $m$, i.e., a lower compression ratio $d/m$, admits a larger learning rate, but it requires more per-round communication overhead.

\subsection{The Optimal Clipping Ratio}
\label{sec:opt}

So far, we have treated the clipping level $\tau_{k,t}$, and hence the clipping ratio $\gamma=\tau_{k,t}/\sigma_s$, as an arbitrary positive constant.
As $\gamma\rightarrow\infty$, nothing is clipped and the transmitter is in pure back-off; its scale $c_t^\star$ in \eqref{eq:cstar} is determined by the peak of the block $\bar{\bs}_{k,t}$, because $\alpha_\gamma=1$.
Lowering $\gamma$ clips more aggressively, which increases the clipping distortion $\Ecl$ and reduces the channel noise $\Ech$ in \cref{prop:moments}.
To find the optimal clipping ratio, we extract the $\gamma$-dependent terms from \eqref{eq:convrate}, i.e., $2\Delta_\textrm{cl} + \Delta_\textrm{ch}$.
Then, denoting $\snr=KP_\textrm{pk}/(N_0\mathbb{E}[\max_{k}h_{k,t}^{-2}])$, we have
\begin{equation}\label{eq:Jsplit}
    2\Delta_{\mathrm{cl}}+\Delta_{\mathrm{ch}}
    =\frac{d}{m}\Big(J(\gamma;\snr)+\frac{2A(\gamma)}{\alpha_{\gamma}^{2}}\,\varepsilon_{\mathrm{cl}}\Big),
\end{equation}
where $J(\gamma;\snr)=\frac{2A(\gamma)+\gamma^{2}/\snr}{\alpha_\gamma^{2}}$.

\begin{corollary}[Optimal clipping ratio]
\label{cor:opt}
    Let
    \begin{equation}\label{eq:Psidef}
        \Psi(\gamma)=\frac{\gamma}{4\big(\phi(\gamma)-\gamma\,Q(\gamma)\big)},
    \end{equation}
    which is strictly increasing  on $(0,\infty)$.
    Then, for every $\snr>0$, $J(\cdot;\snr)$ has the unique minimizer
    \begin{equation}\label{eq:mustar}
        \gamma^\star(\snr)=\Psi^{-1}(\snr).
    \end{equation}
    Writing $\frac{m}{d}\big(2\Delta_{\mathrm{cl}}+\Delta_{\mathrm{ch}}\big)=J(\gamma;\snr)+r(\gamma)$ as in \eqref{eq:Jsplit}, clipping at $\gamma^\star$ exceeds the exact minimum of the clipping-and-channel term of \eqref{eq:Jsplit} by at most
    \begin{align}\label{eq:gapbound}
        & J(\gamma^\star;\snr)+\frac{2A(\gamma^\star)}{\alpha_{\gamma^\star}^2}\varepsilon_{\textrm{cl}}
        -\min_\gamma \left(J(\gamma;\snr)+\frac{2A(\gamma)}{\alpha_{\gamma}^2}\varepsilon_{\textrm{cl}}\right)\nonumber \\
        & ~ \le \frac{1}{d\,\alpha_{\gamma^\star}^{2}}\Big(1+\frac{6m}{\bar d-1}\Big)=\mathcal{O}\Big(\frac1d\Big).
    \end{align}
    The proof can be found in Appendix~\ref{app:opt}.
\end{corollary}

The gap \eqref{eq:gapbound} is worth reading numerically: it is below $10^{-7}$ once $(d,m)=(1.1\times10^{7},2^{14})$, and at $\snr=24.5$~dB, about $3\times10^{-6}$ of $J(\gamma^\star)$ itself.
We can obtain $\gamma^\star=\Psi^{-1}(\snr)$ by the following Newton iteration:
\begin{align}\label{eq:newton}
    \gamma_{n+1}
    &=\gamma_n-\frac{\Psi(\gamma_n)-\snr}{\Psi'(\gamma_n)},
\end{align}
where $\Psi'(\gamma)=\phi(\gamma)/(4(\phi(\gamma)-\gamma Q(\gamma))^2)$.

\section{Numerical Results}

In this section, we numerically evaluate the proposed method (\AirGCCD) for various SNRs and datasets. 

\subsection{Experimental Environments}

In our experiments, we assume a wireless network with a PS and $K=20$ EDs. 
Since our focus is on unbiasedness and communication efficiency, we use an AWGN channel, \ie, $h_{k,t}=1$ for all $k$ and $t$.
Multi-antenna channels reduce to the same model under uniform-forcing beamforming, as noted in \cref{sec:system_model}.
Removing fading isolates the effect of the SNR as in \eqref{eq:Jsplit}, which we vary from $-20$ to $20$~dB.

\paragraph*{Dataset}

Five datasets are used: i) CIFAR-10~\cite{krizhevsky2009learning}, ii) CIFAR-100~\cite{krizhevsky2009learning}, iii) SVHN~\cite{netzer2011reading}, iv) Fashion-MNIST~\cite{xiao2017fashion}, and v) Tiny-ImageNet~\cite{tiny-imagenet}.
Unless otherwise specified, we use CIFAR-10 for the numerical evaluations.
For the data partitioning, we use the Dirichlet distribution with a parameter $\vartheta$, which is the standard way of simulating non-IID data in FL: a smaller $\vartheta$ yields a more skewed class distribution across EDs, and a larger $\vartheta$ yields a more balanced one. 
Unless stated otherwise, we set $\vartheta=0.1$.
CIFAR-10 and CIFAR-100 are augmented with a random $32\times32$ crop (4-pixel reflection padding) and a random horizontal flip; the other datasets use no augmentation. The device partition is fixed by a seed, so every scheme sees exactly the same split.

\paragraph*{Implementation details}

The model is a ResNet-18 trained from scratch, with 11,181,642 trainable parameters. 
Each ED performs 40 local SGD steps with a batch size of 32 and a learning rate of 0.01.
Training runs for 200 communication rounds. 
Unless otherwise specified, the proposed method uses $m=16{,}384$ channel uses per round.
Experiments are run on two NVIDIA RTX A6000 GPUs, under Python 3.9 and PyTorch 2.8.0 with CUDA 12.8.

\paragraph*{Baselines}

For numerical evaluation, we compare the proposed method with the following baseline schemes.
None of them clips their transmitted block; each is scaled so that its own peak meets $P_{\mathrm{pk}}$, \ie, it backs off without clipping, and all schemes therefore operate under the same peak-power constraint.
\begin{itemize}[leftmargin=*]
    \item \textbf{AirFL~\cite{yang2020federated}:} A foundational approach that uses the FedAvg algorithm~\cite{pmlr-v54-mcmahan17a} with AirComp aggregation of the full model updates. This occupies $d$ uplink channel uses per round.  
    \item \textbf{AirFL-LoRA~\cite{kuo2024federated,10763424}:} Low-rank adaptation (LoRA) reduces the communication overhead by adding low-rank adapters to the layers, so that only the adapter is trained and transmitted. We use two rank parameters of $8$ and $16$, which leave 291,560 and 583,120 trainable parameters, that is, 2.6 and 5.2 percent of the full model.
    \item \textbf{AirFL-ZO~\cite{qin2023federated,neto2024communication,mhanna2024rendering}:} Zeroth-order optimization replaces the update by $m$ directional derivatives along shared random directions, transmitting only $m$ coefficients. 
    \item \textbf{AirFL-Sparse:} Top-10\% magnitude sparsification of the update, which occupies $\frac{d}{10}$ uplink channel uses per round.
    \item \textbf{AirFL-Sketch (FedZOE)~\cite{jang2024fed}:} A sketched AirComp-FL scheme that projects the update onto the rows of a structured orthogonal transform (SRHT). Like \AirGCCD, it costs $\mathcal{O}(d\log d)$ and occupies $m$ channel uses per round.
\end{itemize}

We note that our work differs from~\cite{jang2024fed} by analyzing the effect of the clipping operator, optimizing the clipping level, and showing unbiasedness of the aggregation. 

\subsection{Effect of the Clipping Ratio in \AirGCCD}

\begin{figure}[t]
    \centering
    \includegraphics[width=0.99\linewidth]{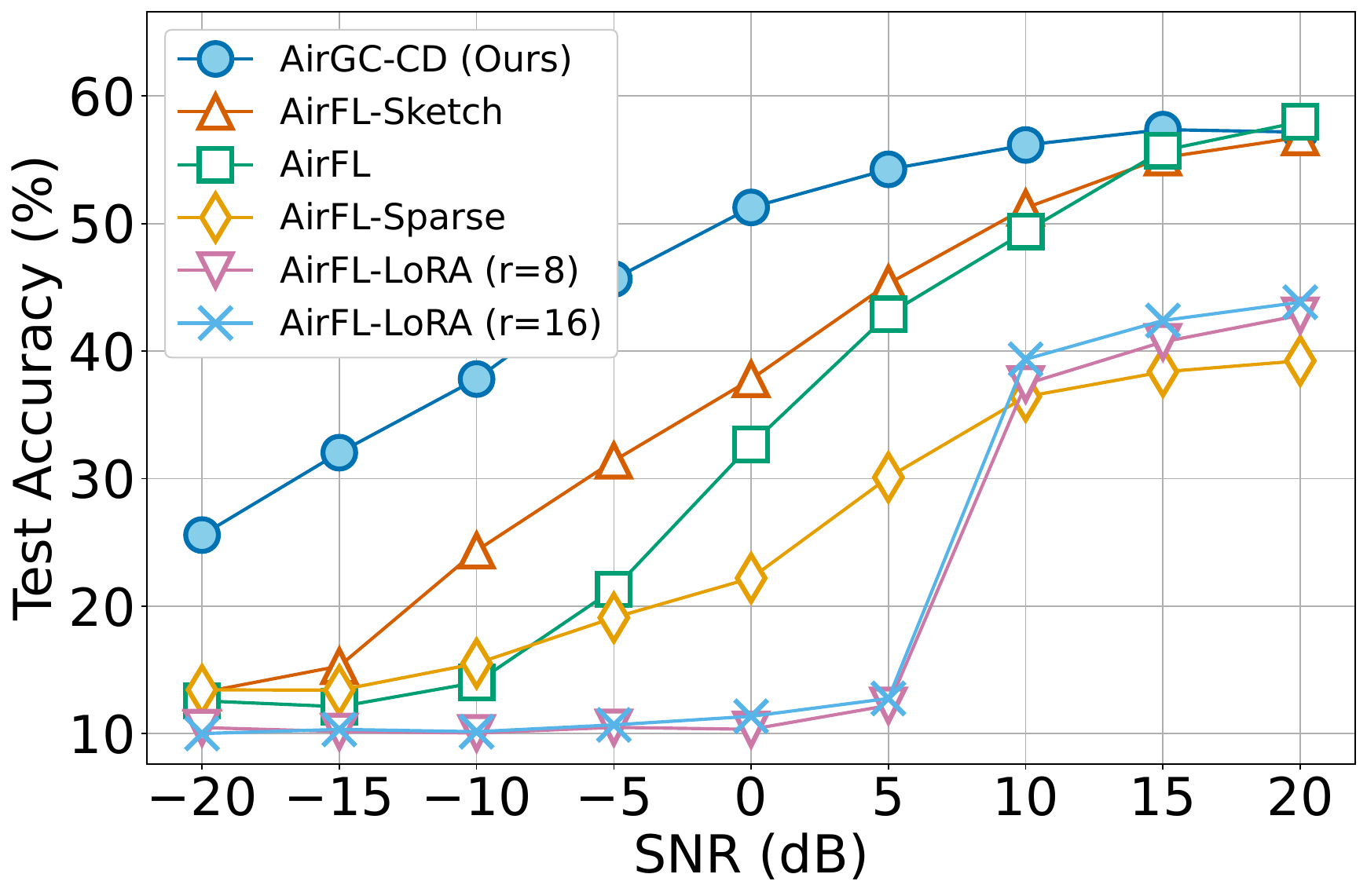}
    \caption{Test accuracy of the proposed method and baselines for various SNRs.}\label{fig:E1_acc_vs_cifar10}
\end{figure}

In \cref{fig:E1_acc_vs_cifar10}, we measure the test accuracy of the proposed method and baselines for SNRs from $-20$~dB to 20~dB with the CIFAR-10 dataset.
As depicted in the figure, the proposed method outperforms the baselines, especially in the low-SNR regime. 
\AirGCCD optimizes the clipping level for better convergence, whereas all the baselines transmit without a clipping operator.
At high SNR, however, \AirGCCD performs similarly to AirFL and AirFL-Sketch, because $\gamma^\star$ in \eqref{eq:mustar} is large and little is clipped.

\begin{figure}[t]
    \centering
    \includegraphics[width=1.0\linewidth]{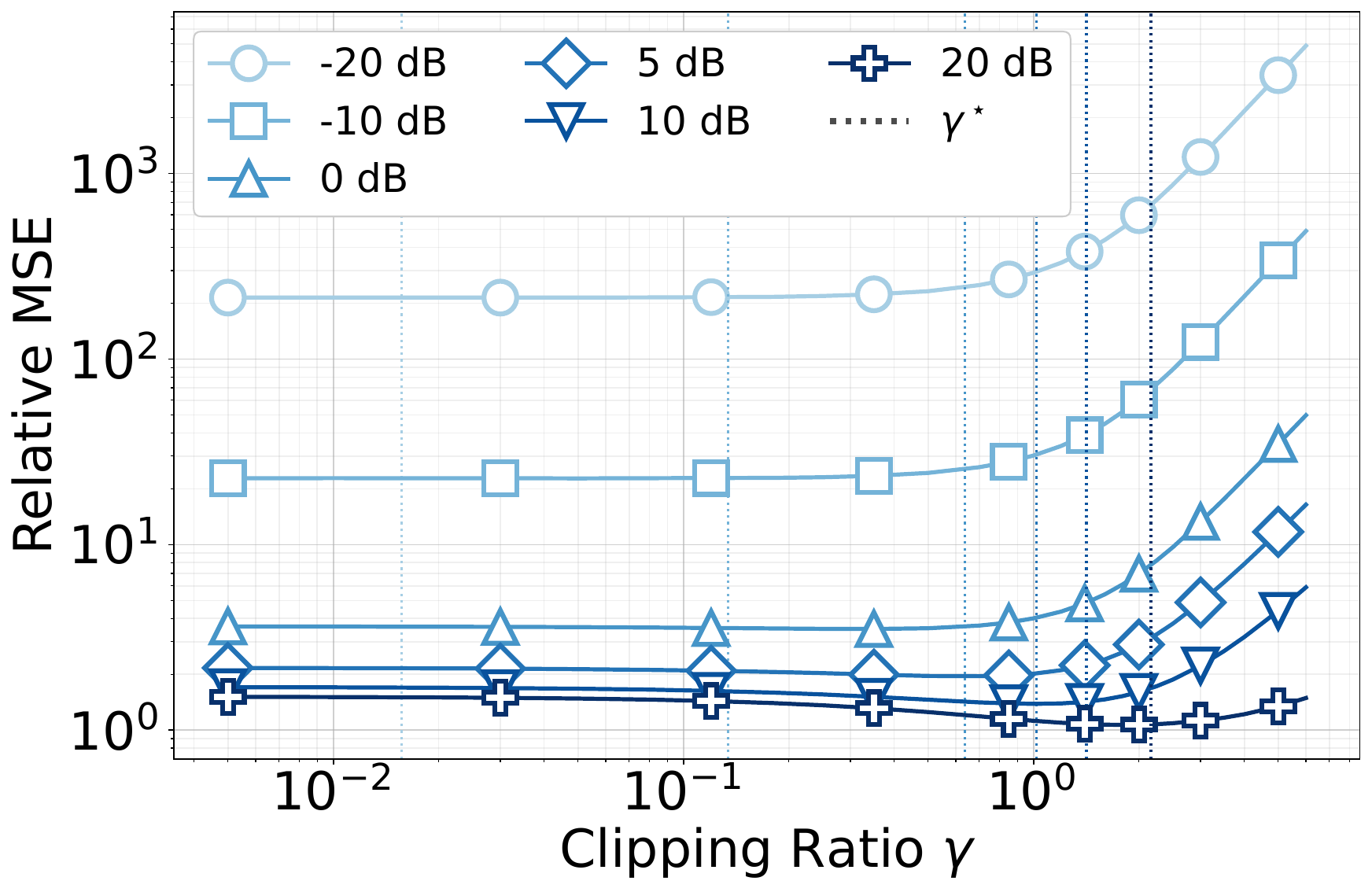}
    \caption{Relative MSE ($\mathbb{E}\big[\Vert\Delta\widehat{\bw}_{G,t}-\tfrac1K\textstyle\sum_k\Delta\bw_{k,t}\Vert^2\big/\Vert\tfrac1K\textstyle\sum_k\Delta\bw_{k,t}\Vert^2\big]$) of update aggregation using \AirGCCD for various clipping ratios $\gamma$ and SNRs. The dotted vertical lines denote the clipping ratio $\gamma^\star$ in \eqref{eq:mustar}.}
    \label{fig:Figure_E4_rel_mse_cifar10_a1}
\end{figure}

To understand where this gain comes from, we depict the relative MSE of $\Delta\widehat{\bw}_{G,t}$ in \cref{fig:Figure_E4_rel_mse_cifar10_a1}. 
The relative MSE gradually increases as the SNR decreases; more importantly, the optimal ratio $\gamma^\star$ decreases because \AirGCCD prefers clipping to back-off for reducing the channel-noise term in \eqref{eq:convrate}.
The measured minimizer of the relative MSE agrees with $\gamma^\star$ of \eqref{eq:mustar} at every SNR, although $\gamma^\star$ is derived from the bound.

\begin{figure}[t]
    \centering
    \includegraphics[width=1.0\linewidth]{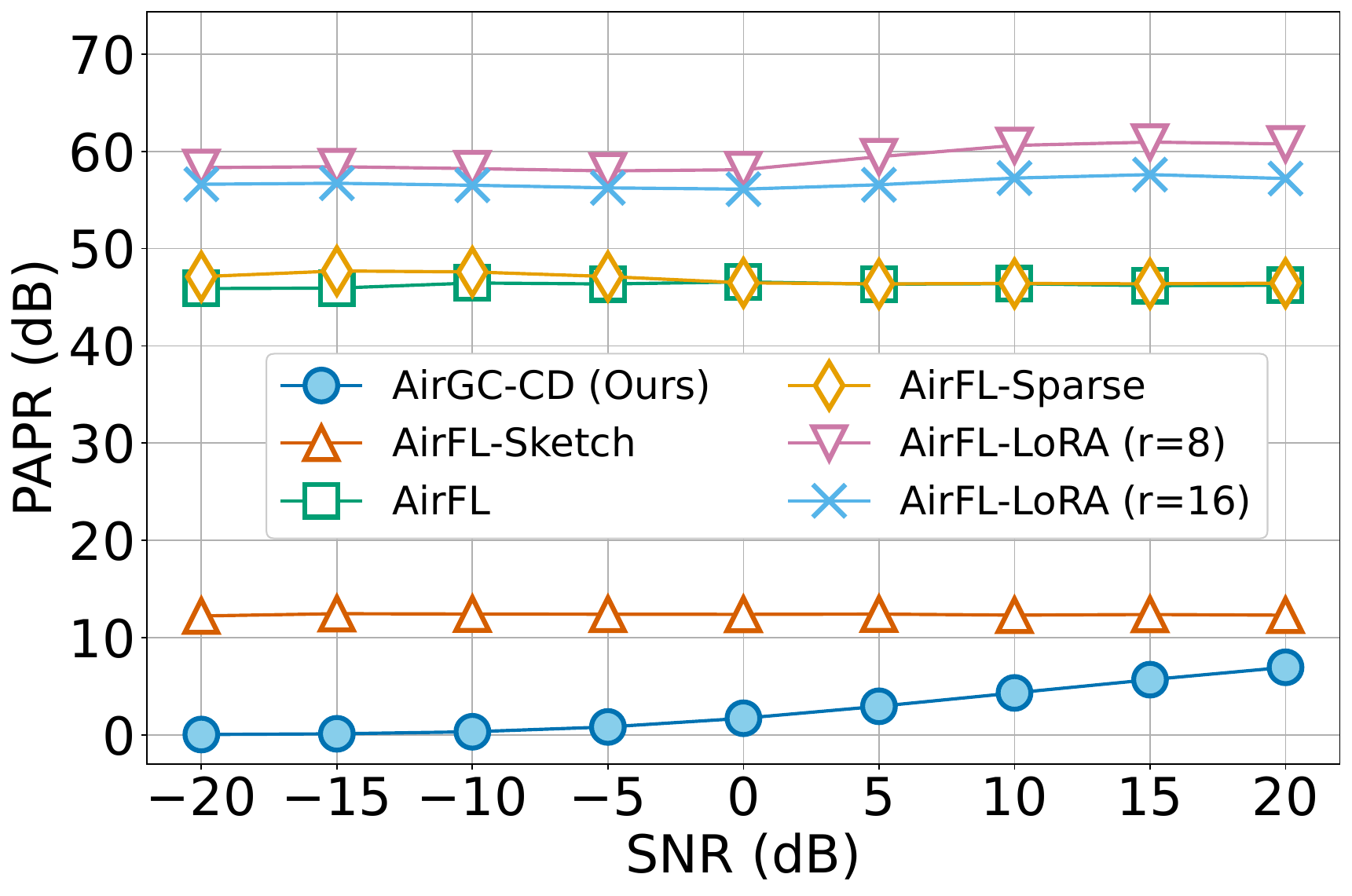}
    \caption{PAPR of the transmitted updates.}
    \label{fig:papr_vs_snr_cifar10}
\end{figure}

In \cref{fig:papr_vs_snr_cifar10}, we depict the PAPR of the transmitted updates for \AirGCCD and baselines. 
Each baseline's PAPR is almost constant over SNRs because no clipping operator is applied. 
Also, AirFL-Sparse has a similar PAPR to the naive AirFL, while AirFL-LoRA has a higher PAPR value.
Meanwhile, \AirGCCD optimizes the clipping ratio for a given SNR; thus, its PAPR is almost $0$~dB in the low-SNR regime, as the optimal clipping ratio $\gamma^\star$ is small, as shown in \cref{fig:Figure_E4_rel_mse_cifar10_a1}.
The PAPR of \AirGCCD increases with SNR due to the larger clipping ratio $\gamma^\star$.

\subsection{System Overhead and Convergence}

\begin{figure}[t]
    \centering
    \subfloat[SNR=0~dB\label{Figure_E2b_acc_vs_m_cifar10_a}]{\includegraphics[width=0.48\linewidth]{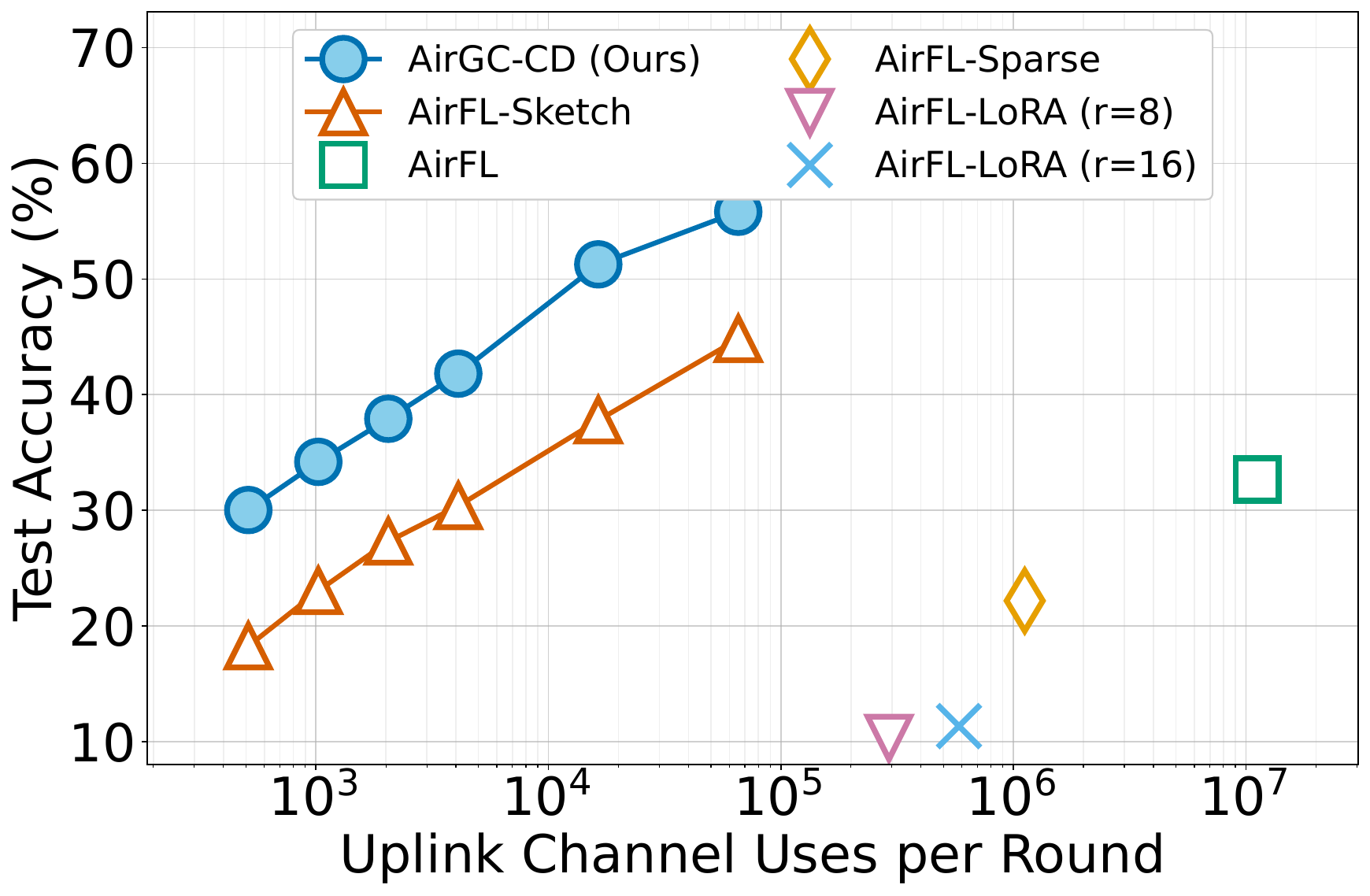}}
    \hfill
    \subfloat[SNR=20~dB\label{Figure_E2b_acc_vs_m_cifar10_b}]{\includegraphics[width=0.48\linewidth]{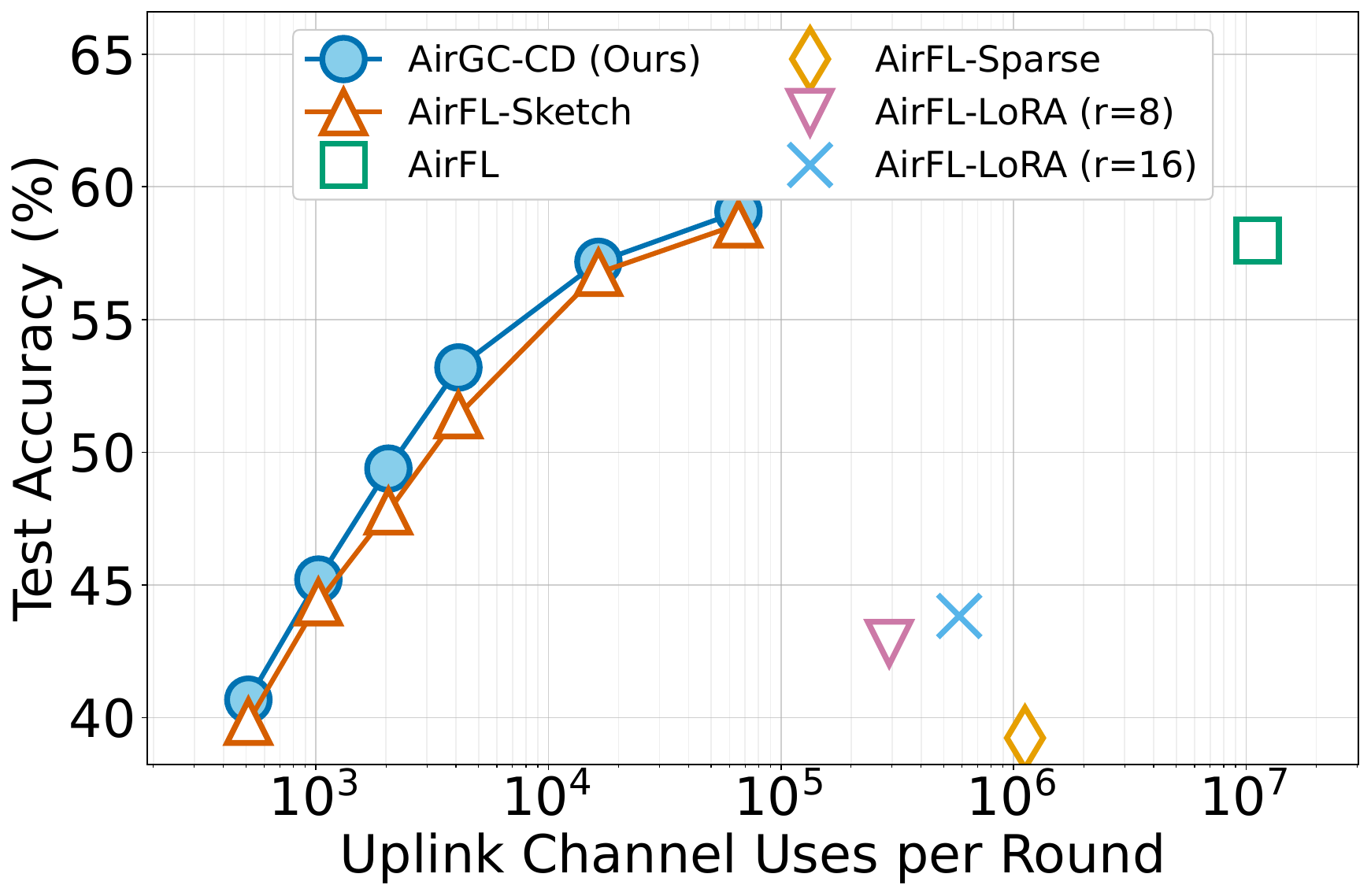}}

    \caption{Test accuracy evaluation with the uplink channel uses per round of \AirGCCD and baselines. The SNR is assumed to be 0~dB and 20~dB for (a) and (b), respectively. For this evaluation, \AirGCCD and AirFL-Sketch are evaluated for $m\in\{512, 1024, 2048, 4096, 16384, 65536\}$.}
    \label{fig:Figure_E2b_acc_vs_m_cifar10}
\end{figure}

Here, we discuss the system overhead of \AirGCCD.
As we mentioned, \AirGCCD reduces the communication overhead from $d\approx1.1\times10^7$ to $m$ by sketching with $\bPhi_t$.
In \cref{fig:Figure_E2b_acc_vs_m_cifar10}, we compare the test accuracy and communication overhead of AirComp-FL schemes for SNR$\in\{0,20\}$ dB. 
In \cref{Figure_E2b_acc_vs_m_cifar10_a}, the proposed method attains higher accuracy with fewer channel uses than AirFL, AirFL-Sparse, and AirFL-LoRA by virtue of its sketching matrix design. 
Compared to AirFL-Sketch, the proposed method occupies the same number of uplink channel uses because both schemes use sketching matrices of the same dimension. 
Nevertheless, for a fixed $m$, \AirGCCD outperforms AirFL-Sketch because it can clip without introducing bias, whereas AirFL-Sketch must remain unclipped and back off to its peak.
For a higher SNR in \cref{Figure_E2b_acc_vs_m_cifar10_b}, the proposed method still outperforms the baselines; however, the gap narrows, because $\gamma^\star$ is large and little is clipped.

\begin{figure}[t]
    \centering
    \includegraphics[width=1.0\linewidth]{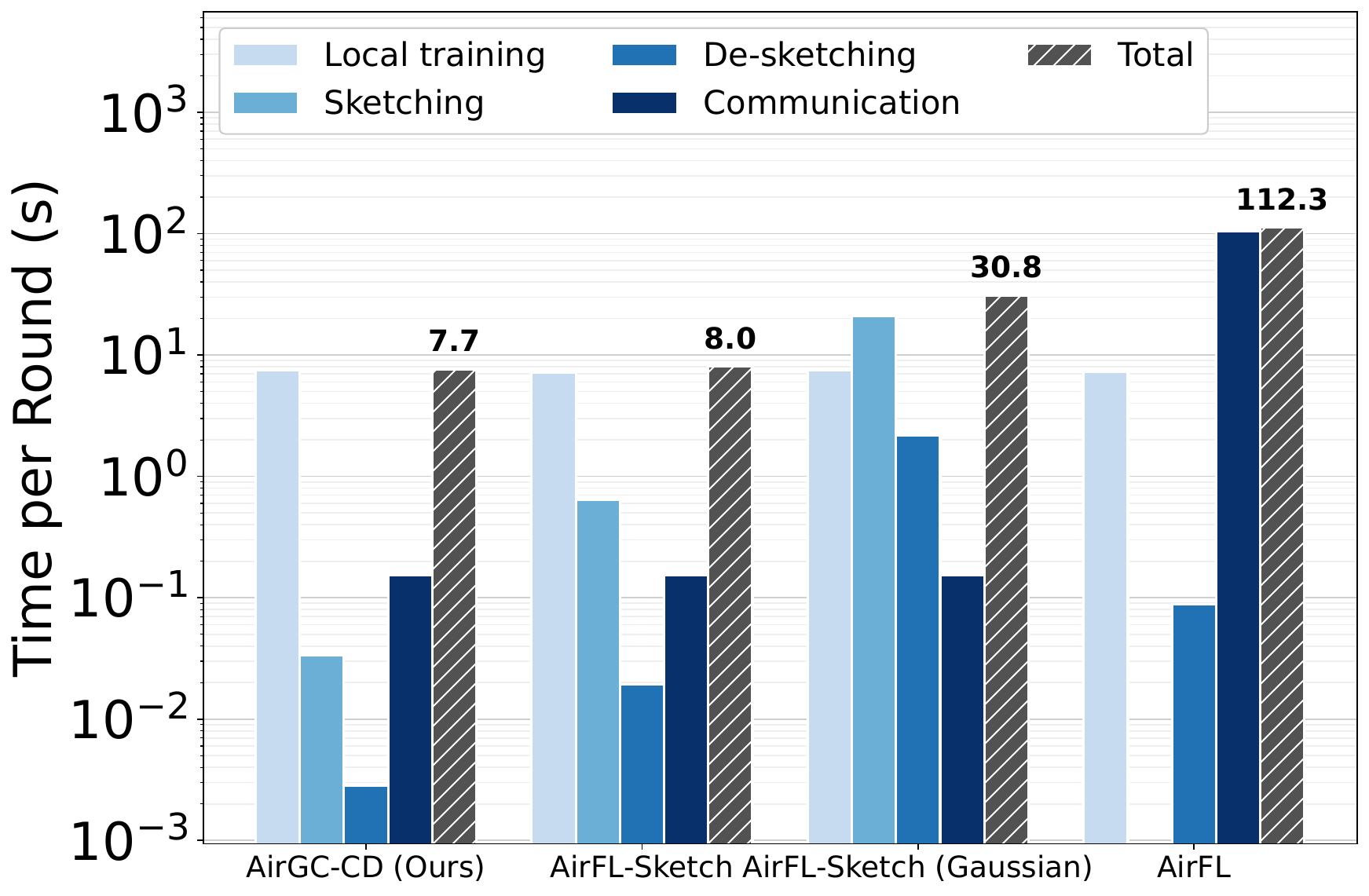}
    \caption{Per-round system overhead: local training, sketching, desketching, and communication.}
    \label{fig:overhead}
\end{figure}

For a more detailed analysis of system overhead, we depict the overhead of i) local training, ii) sketching, iii) desketching, and iv) communication in \cref{fig:overhead}. 
The communication overhead assumes a symbol rate of $10^6$ symbols/s and a downlink rate of 4 Mbps. 
The local-training overhead is identical across schemes.
The total overheads of \AirGCCD and AirFL-Sketch are the smallest, owing to their low sketching and desketching cost. 
Although the SRHT-based sketch of AirFL-Sketch costs about as much as \AirGCCD, it does not guarantee unbiased aggregation when clipping is applied.
On the other hand, the dense Gaussian sketch in AirFL-Sketch (Gaussian) guarantees unbiased aggregation but incurs a substantial sketching overhead. 
Thus, the proposed method is the only approach guaranteeing unbiased clipped aggregation while having an $\mathcal{O}(d\log d)$ complexity sketch. 

\begin{figure}[t]
    \centering
    \subfloat[Test accuracy versus communication rounds for SNR=0~dB.\label{fig:testacc_overhead_convergence_a}]{\includegraphics[width=0.48\linewidth]{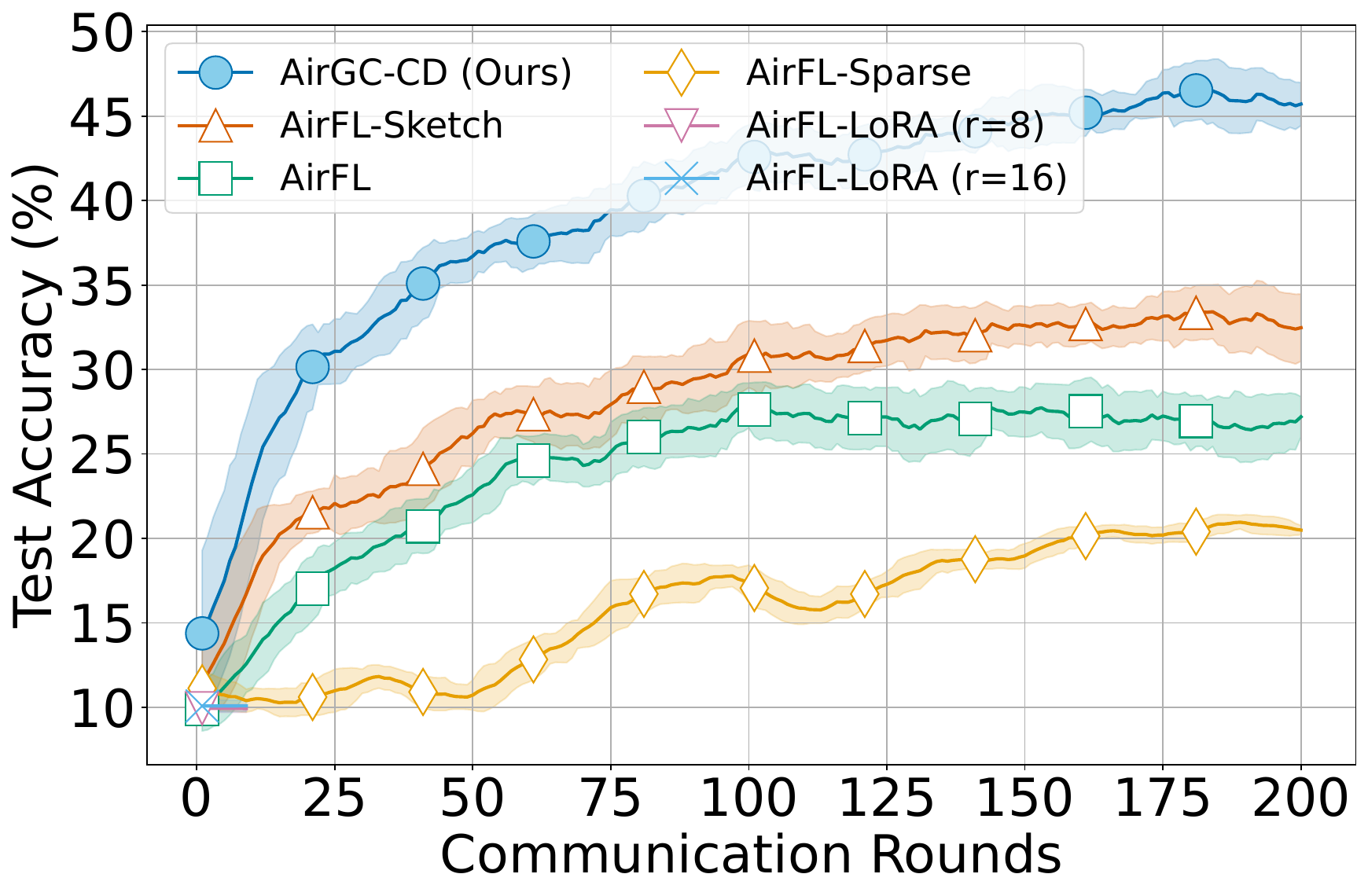}}
    \hfill
    \subfloat[Test accuracy versus communication rounds for SNR=10~dB.\label{fig:testacc_overhead_convergence_b}]{\includegraphics[width=0.48\linewidth]{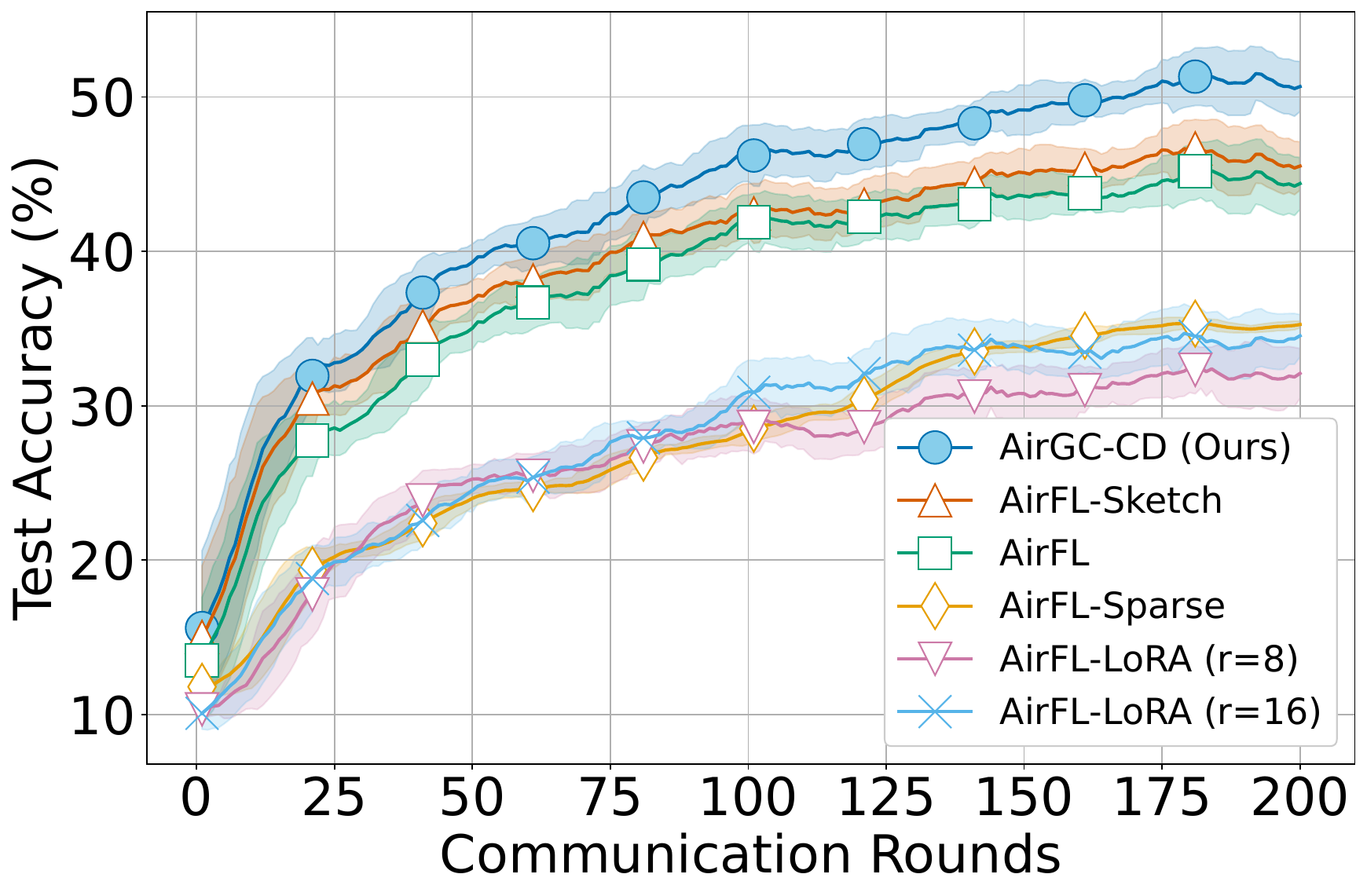}}
    \\
    \subfloat[Test accuracy versus total communication overhead for SNR=0~dB.\label{fig:testacc_overhead_convergence_c}]{\includegraphics[width=0.48\linewidth]{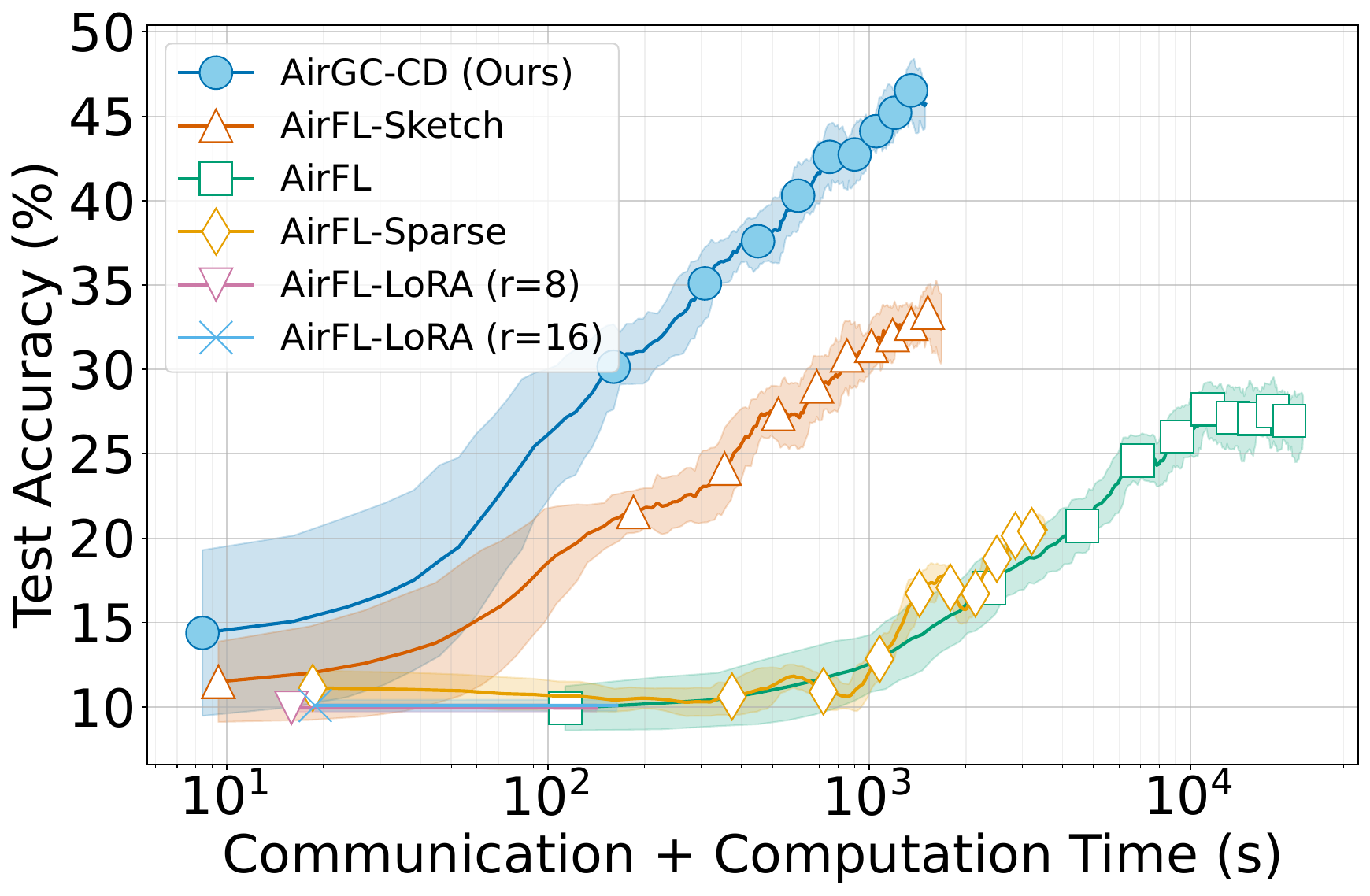}}
    \hfill
    \subfloat[Test accuracy versus total communication overhead for SNR=10~dB.\!\!\label{fig:testacc_overhead_convergence_d}]{\includegraphics[width=0.48\linewidth]{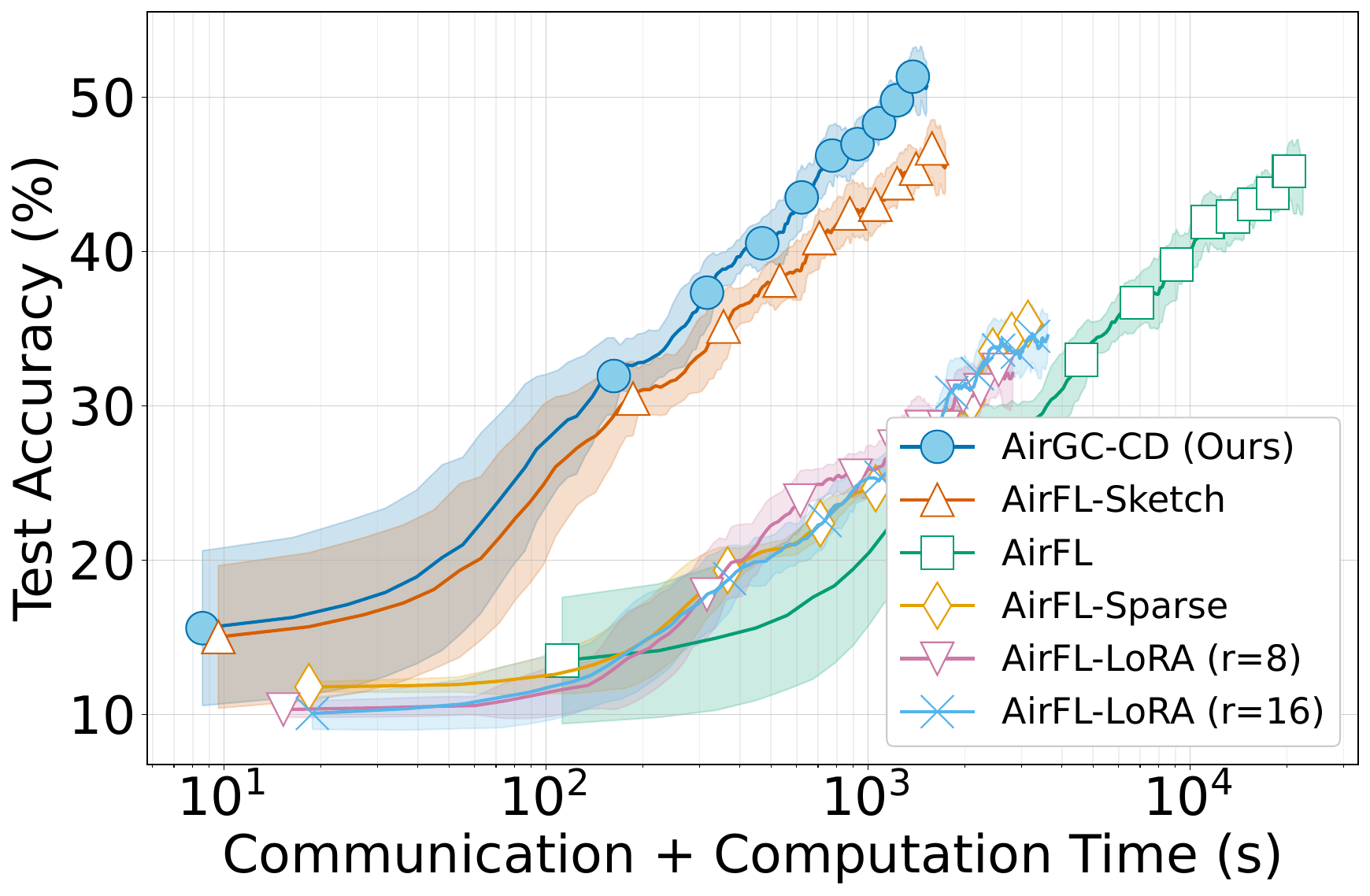}}
    \caption{The convergence of \AirGCCD and baselines for SNR$\in\{0,10\}$ dB. Subfigures (a) and (b) depict the test accuracy versus communication rounds, and subfigures (c) and (d) depict the test accuracy versus total communication overhead. AirFL-LoRA is omitted from (a) and (c) because it diverges at $0$~dB.}
    \label{fig:testacc_overhead_convergence}
\end{figure}

To further analyze the convergence and overhead of \AirGCCD, we depict the test accuracy of \AirGCCD versus communication rounds and total communication overhead in \cref{fig:testacc_overhead_convergence}.
In \cref{fig:testacc_overhead_convergence_a,fig:testacc_overhead_convergence_b}, \AirGCCD attains a higher test accuracy than baselines at every communication round, i.e., it converges faster.
The gain from clipping optimization is larger at low SNR.
In \cref{fig:testacc_overhead_convergence_c,fig:testacc_overhead_convergence_d}, since \AirGCCD also has the smallest per-round overhead, its advantage widens when accuracy is measured against total overhead.

\subsection{Sketch Dimension and Learning Rate}

\begin{figure}[t]
    \centering
    \includegraphics[width=1.0\linewidth]{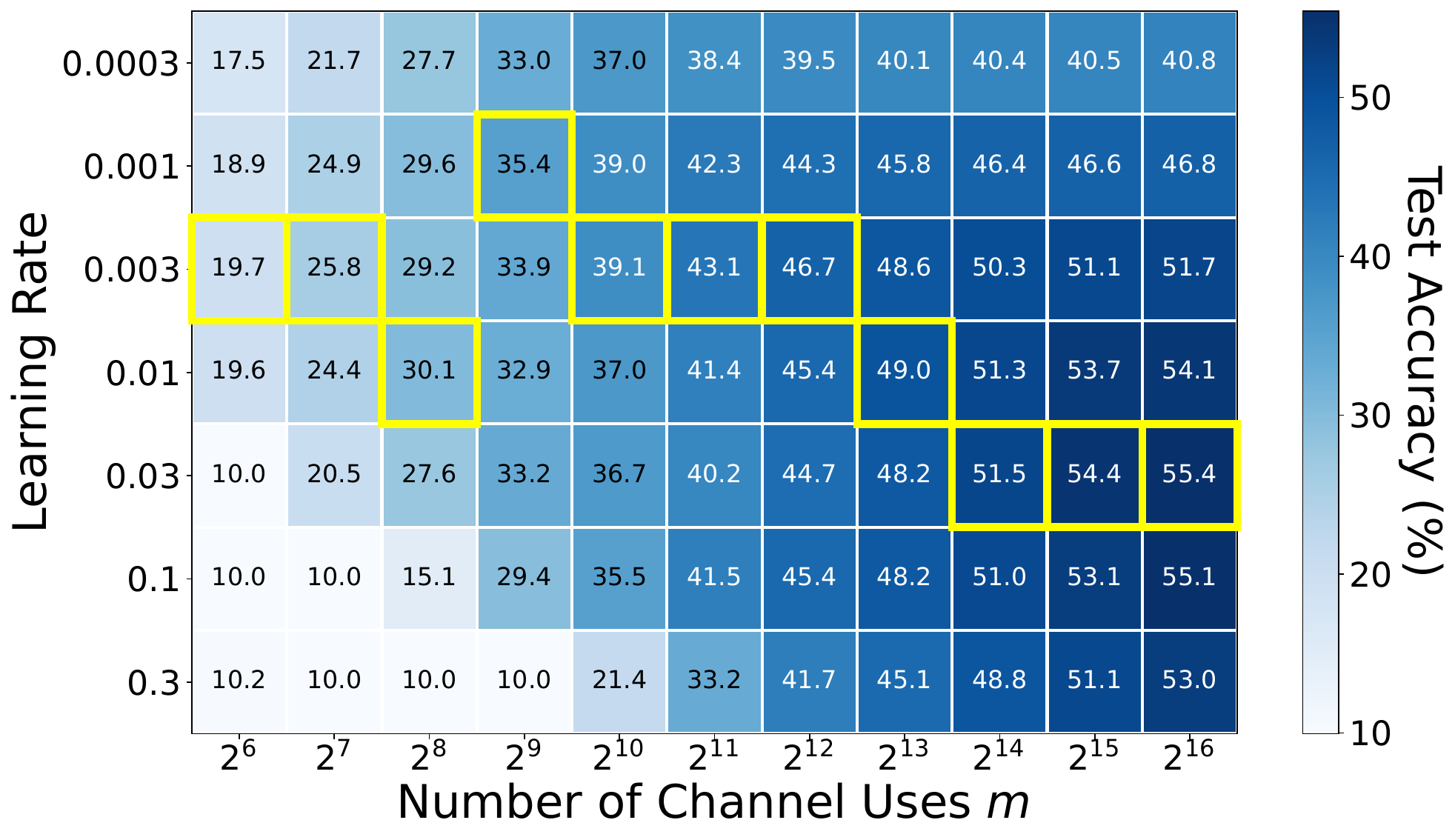}
    \caption{The test accuracy of the trained model for various learning rates $\eta$ and sketch dimensions $m$. The best test accuracy for each $m$ is highlighted. }
    \label{fig:lr_vs_m}
\end{figure}

In this subsection, we analyze the effect of $m$ on the admissible learning rate $\eta$ defined in \cref{thm:conv}, i.e., $\eta\ =\frac{1}{\sqrt{TI}}\le\ \tfrac{1}{\beta\big(1+2I\Delta_{\mathrm{sk}}\big)}$, where $\Delta_\mathrm{sk}$ decreases with $m$, so a larger $m$ admits a larger $\eta$.
To verify this relationship, we depict the test accuracy for various learning rates and sketching dimensions $m$ in \cref{fig:lr_vs_m}.
As depicted in the figure, with small $m$, e.g., $2^6$ and $2^7$, training diverges at a learning rate of $0.1$. 
As $m$ increases, however, the learning rate of $0.1$ converges better than the lower learning rate of $0.003$, which is the best for $m\in\{2^6,2^7\}$.
These results confirm that a larger $m$ admits a larger learning rate, as in \cref{thm:conv}.

\subsection{Various Datasets}

\begin{figure}[t]
    \centering
    \subfloat[CIFAR-100.\label{subfig:datasets_a}]{\includegraphics[width=0.48\linewidth]{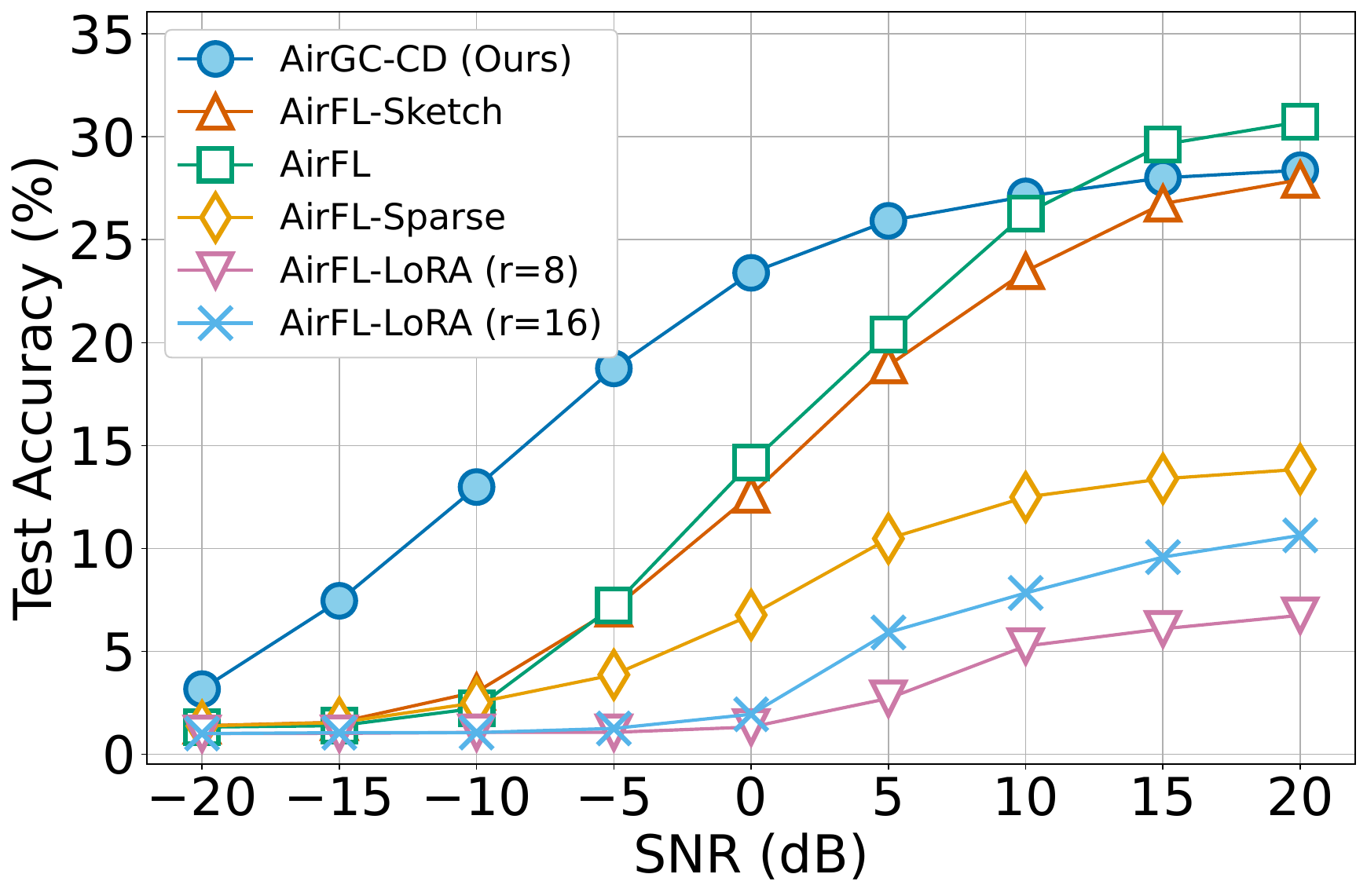}}
    \hfill
    \subfloat[SVHN.\label{subfig:datasets_b}]{\includegraphics[width=0.48\linewidth]{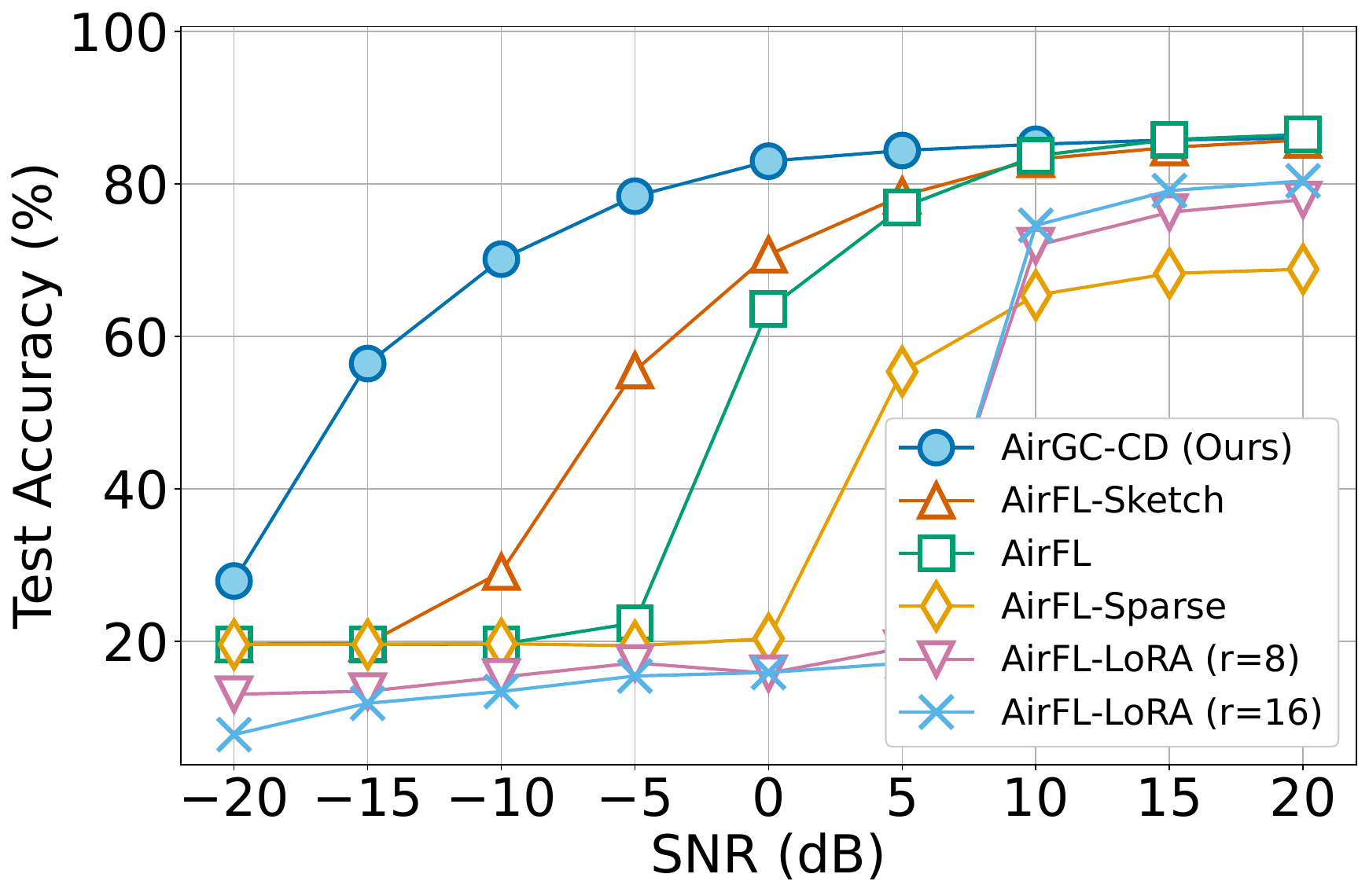}}\\
    \subfloat[Fashion-MNIST.\label{subfig:datasets_c}]{\includegraphics[width=0.48\linewidth]{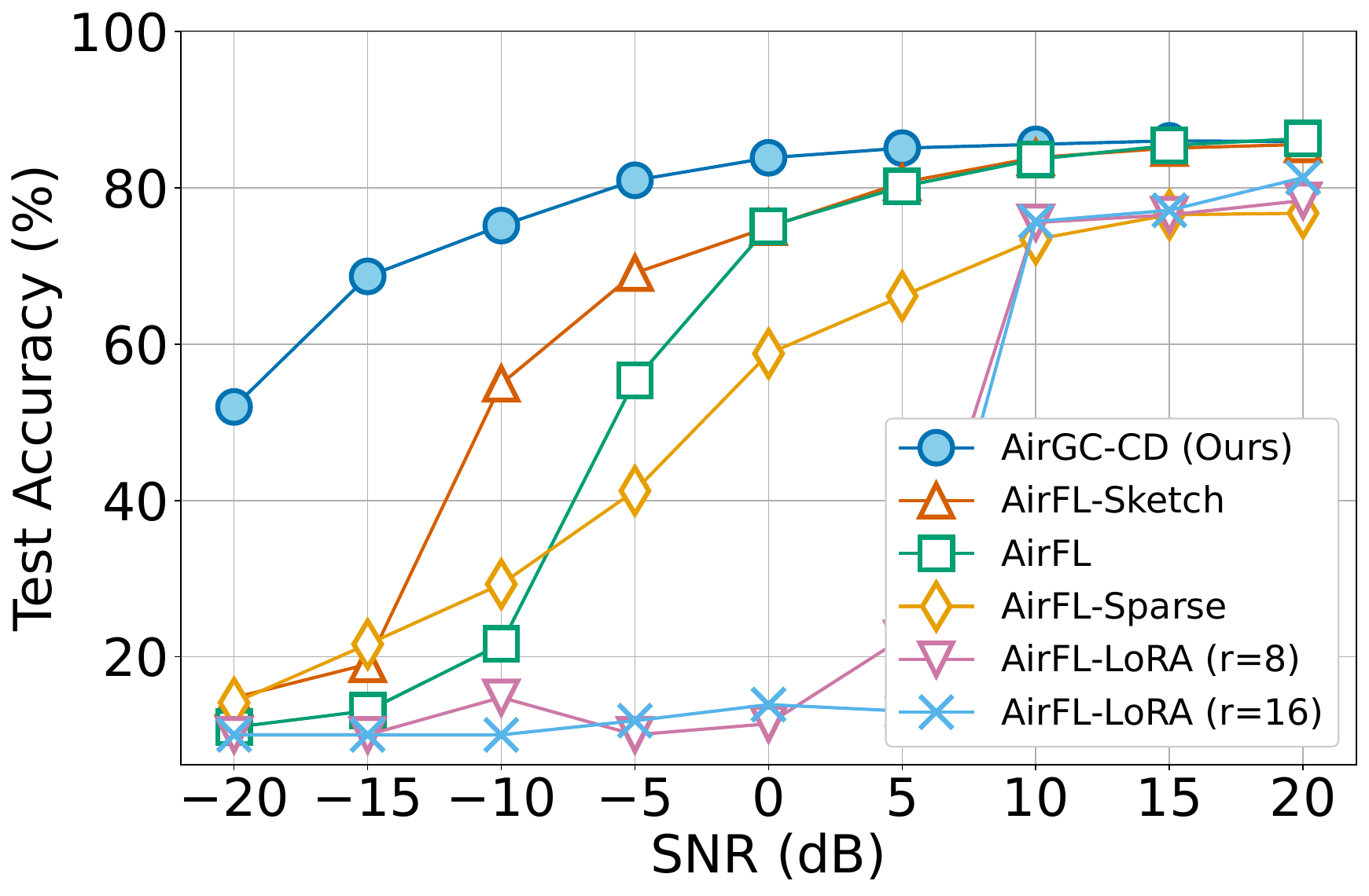}}
    \hfill
    \subfloat[Tiny-ImageNet.\label{subfig:datasets_d}]{\includegraphics[width=0.48\linewidth]{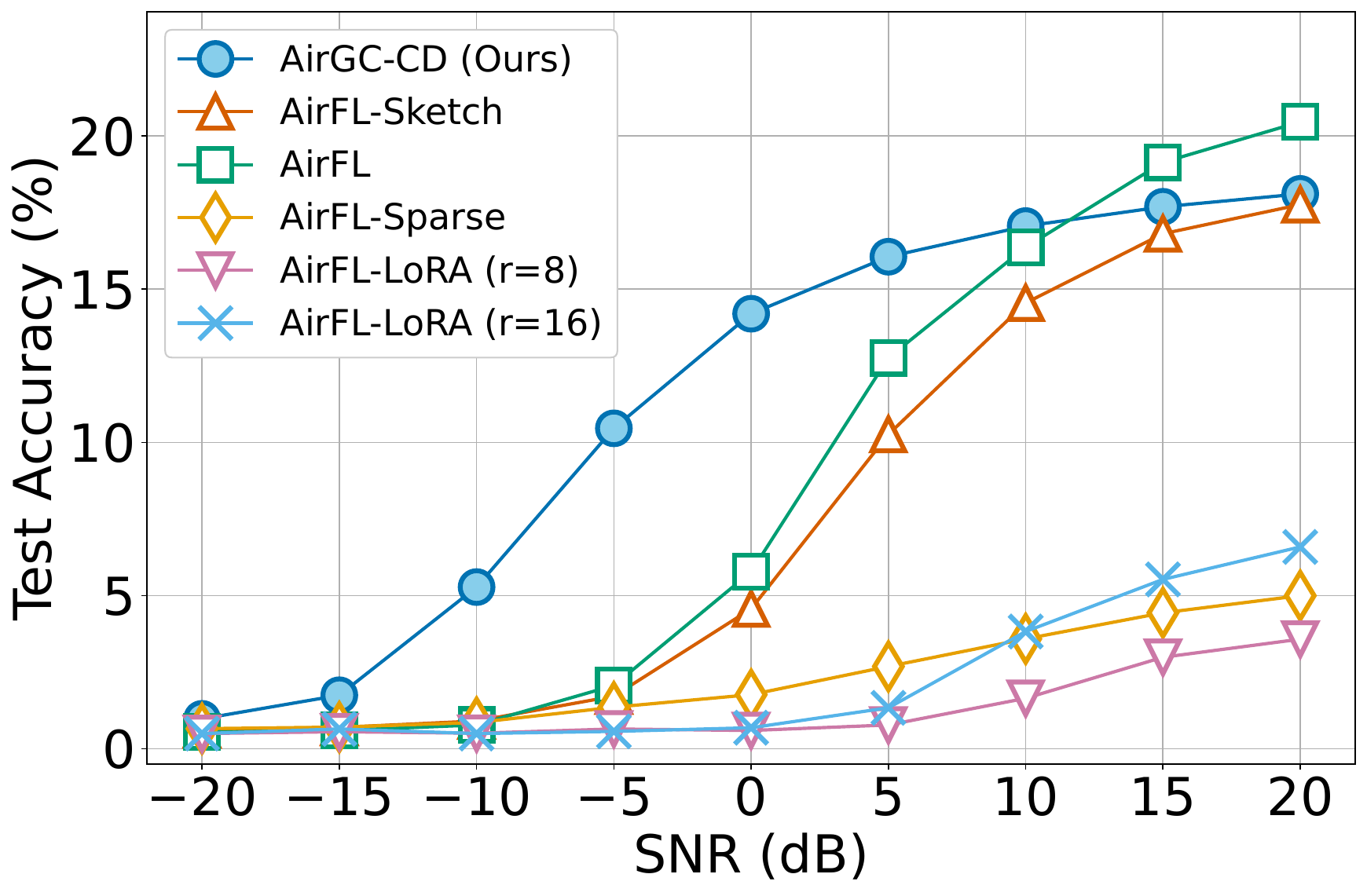}}
    \caption{Test accuracy of \AirGCCD and baselines for various datasets. }
    \label{fig:datasets}
\end{figure}

This subsection evaluates \AirGCCD for various datasets, including CIFAR-100, SVHN, Fashion-MNIST, and Tiny-ImageNet.
In \cref{fig:datasets}, we show the test accuracy of \AirGCCD and baselines for various SNRs and datasets. 
\AirGCCD outperforms the baselines in most settings.
However, for harder tasks in \cref{subfig:datasets_a,subfig:datasets_d}, \AirGCCD and AirFL-Sketch fall below uncompressed AirFL in the high-SNR regime. 
This is because the sketching dimension, $m=16{,}384$, is small relative to the model size in this experiment.
As observed for CIFAR-10 in \cref{fig:Figure_E2b_acc_vs_m_cifar10}, the gap is expected to close as $m$ increases.

\section{Conclusion}

This paper has addressed over-the-air FL under a peak-power constraint, where each ED must choose between clipping, whose distortion biases the aggregate, and back-off, which lowers the received SNR, and where the uncompressed uplink spends one channel use per model parameter.
We propose \AirGCCD, which precodes each local update with a partial Gaussian-circulant sketch drawn from a shared seed and clips the precoded block before transmission.
The central idea is to choose the precoder for the \textit{distribution} of its output rather than for the flatness of the block: every precoded coordinate is exactly Gaussian whatever the update, so the clipping distortion depends on the clipping ratio alone and is inverted in the mean by a single Bussgang gain that every ED and the PS compute without knowing the updates.
On this basis, we proved that the aggregate is exactly unbiased for every update and every clipping ratio, that the same operator compresses the uplink from $d$ to $m$ channel uses at $\mathcal{O}(d\log d)$ cost while leaving the desketched channel noise equal to that of uncompressed AirComp-FL, and that learning reaches a stationary point at the rate $\mathcal{O}(1/\sqrt{T})$ with no bias floor, the compression ratio being the only distortion that tightens the admissible learning rate.
The analysis also reduces the choice between clipping and back-off to a single scalar: the clipping loss decays super-exponentially in the clipping ratio while the channel noise grows only quadratically, and the optimal ratio $\gamma^\star=\Psi^{-1}(\snr)$ is unique, computable by a Newton iteration, and within $\mathcal{O}(1/d)$ of the exact minimum of the clipping-and-channel term of the bound.

\bibliographystyle{IEEEtran}
\bibliography{main}

\appendices

\section{Proof of \cref{lem:be}}
\label{app:be}

In this appendix, for notational brevity, we drop the subscripts $k$ and $t$, \ie, $\bar\bs=\bar\bs_{k,t}$ and $\Delta\bw=\Delta\bw_{k,t}$.
We define the padded ($\bJ$), sign-flipped ($\bD$) update as 
\begin{equation}\label{eq:be_z}
    \bz = \bD\bJ\Delta\bw\in\mathbb{R}^{\bar d},
\end{equation}
where $\Vert\bz\Vert=\Vert\Delta\bw\Vert$; note that $[\bz]_j=0$ for $j>d$.
Let $\Omega=\{\omega_1,\dots,\omega_m\}\subset[\bar d]$ denote the sampled rows; by the definition of the circulant matrix in \eqref{eq:circ}, the $i$-th coordinate of $\bar\bs\in\mathbb{R}^m$ is
\begin{equation}\label{eq:be_inner}
    [\bar\bs]_i=\frac{\big\langle\bc^{(\omega_i)},\bz\big\rangle}{\sqrt m},
    ~~~ i\in[m].
\end{equation}

Conditioned on $(\bD,\bP_\Omega)$, the coordinate in \eqref{eq:be_inner} is a linear function of $\bc\sim\mathcal{N}(\mathbf{0},\bI_{\bar d})$ and is therefore Gaussian:
\begin{equation}\label{eq:be_claim1}
    [\bar\bs]_i\ \sim\ \mathcal{N}\left(0,\frac{\norm{\bz}_2^{2}}{m}\right) = \mathcal{N}\left(0,\frac{\Vert\Delta\bw\Vert_2^{2}}{m}\right),
\end{equation}
which completes the claim.

\section{Proof of \cref{lem:bussgang}}
\label{app:gauss}

Consider $\varsigma\sim\mathcal{N}(0,\sigma_s^2)$, whose probability density function is written as $p(x) = \phi(x/\sigma_s)/\sigma_s$, and recall that $\tau=\gamma\sigma_s$.
Before proving the claims, we list two properties of the standard normal density that follow from $d\phi(x)/dx=-x\phi(x)$:
\begin{equation}\label{eq:normal_property}
    \begin{cases}
        \int_\gamma^\infty x \phi(x) dx = \phi(\gamma), \\
        \int_{-\gamma}^\gamma x^2 \phi(x) dx= -2\gamma\phi(\gamma) +\alpha_\gamma.
    \end{cases}
\end{equation}
Then, we have
\begin{align}
    &\mathbb{E}[\varsigma\clip_\tau(\varsigma)] = \int_{-\infty}^\infty \frac{\phi(x/\sigma_s)}{\sigma_s}x\clip_\tau(x)dx \\
    &~~= \sigma_s^2\left( \int_{-\gamma}^\gamma x^2\phi(x)dx+2\int_\gamma^\infty \gamma x \phi(x) dx\right)\nonumber \\
    &~~ \overset{(a)}{=} \sigma_s^2\left(\alpha_\gamma - 2\gamma\phi(\gamma)+2\gamma\phi(\gamma)\right) = \alpha_\gamma\sigma_s^2,\nonumber
\end{align}
where the equality (a) holds from \eqref{eq:normal_property}. 
Similarly, we have
\begin{align}
    & \mathbb{E}[\clip_\tau(\varsigma)^2]= \int_{-\infty}^\infty \frac{\phi(x/\sigma_s)}{\sigma_s}\clip^2_\tau(x)dx\\
    \nonumber & ~~~= \sigma_s^2\left(\int_{-\gamma}^\gamma x^2\phi(x)dx + 2\gamma^2\int_\gamma^\infty\phi(x)dx \right)\\
    \nonumber & ~~~ = \sigma_s^2(\alpha_\gamma - 2\gamma\phi(\gamma) +2\gamma^2Q(\gamma)) = \omega(\gamma)\sigma_s^2.
\end{align}
Finally, the last claim, on $\mathbb{E}[f(\varsigma)^2]$, follows from
\begin{align}
    & \mathbb{E}[f(\varsigma)^2] = \mathbb{E}[(\clip_\tau(\varsigma) - \alpha_\gamma\varsigma)^2] \\
    & ~~~ = \omega(\gamma)\sigma_s^2 - 2\alpha^2_\gamma\sigma_s^2 + \alpha_\gamma^2\sigma_s^2 = A(\gamma)\sigma_s^2, \nonumber
\end{align}
which completes the claim.

\section{Proof of \cref{lem:sketch}}
\label{app:sketch}

For $\bu\in\mathbb{R}^{d}$,  we write $\bz=\bD\bJ\bu\in\mathbb{R}^{\bar d}$, so that $\Vert\bz\Vert_2^2=\Vert\bu\Vert_2^2$and
\begin{equation}\label{eq:be_dmean}
    \mathbb{E}_{\bd}\big[[\bz]_p[\bz]_{p+\ell}\big]=0,
\end{equation}
for $\ell\bmod\bar d \neq 0$, since the indices $p$ and $p+\ell$ are distinct and the signs are independent with zero mean.

\paragraph*{Claim (i)}
From \eqref{eq:sketchmat}, we have
\begin{equation}\label{eq:sk_gram}
    \bPhi_t^\top\bPhi_t=\frac{1}{m}\bJ^\top\bD\bC^\top\bP_\Omega^\top\bP_\Omega\bC\bD\bJ.
\end{equation}
The matrix $\bP_\Omega^\top\bP_\Omega$ is diagonal with  $[\bP_\Omega^\top\bP_\Omega]_{i,i}=\mathbbm{1}_{\{i\in\Omega\}}$, and uniform sampling without replacement retains each index with probability $m/\bar d$, so $\mathbb{E}_\Omega\big[\bP_\Omega^\top\bP_\Omega\big]=\frac{m}{\bar d}\bI_{\bar d}$.
Taking the expectation of \eqref{eq:sk_gram} over $\Omega$ and then over $\bc$, and using $\mathbb{E}_{\bc}[\bC^\top\bC]=\bar d\bI_{\bar d}$, we have
\begin{equation}\label{eq:sk_claim1}
    \mathbb{E}\big[\bPhi_t^\top\bPhi_t\big]
    =\frac{1}{m}\cdot\frac{m}{\bar d}\cdot\bJ^\top\bD\,\mathbb{E}_{\bc}\big[\bC^\top\bC\big]\bD\bJ
    =\bJ^\top\bJ=\bI_{d},
\end{equation}
which completes the first claim.

\paragraph*{Claim (ii)}

In this proof, it is convenient to define the $\bar{d}$-dimensional vector,
\begin{equation}\label{eq:sk_v}
    \bv=\frac{1}{m}\bC^\top\bP_\Omega^\top\bP_\Omega\bC\bz\in\mathbb{R}^{\bar d},
\end{equation}
where $\bPhi_t^\top\bPhi_t\bu=\bJ^\top\bD\bv$.
Then, by claim (i), $\mathbb{E}_{\bc,\Omega}[[\bv]_j\mid\bD]=[\bz]_j$; since $\bJ^\top$ retains only $j\in[d]$ and $\bPhi_t^\top\bPhi_t\bu=\bJ^\top\bD\bv$ and $\bu=\bJ^\top\bD\bz$, we have 
\begin{align}\label{eq:sk_err}
    \mathbb{E}\big\Vert(\bPhi_t^\top\bPhi_t-\bI_d)\bu\big\Vert^{2}
    &= \mathbb{E}\Vert \bJ^\top \bD(\bv-\bz)\Vert_2^2  = \mathbb{E}\Vert[\bv-\bz]_{[d]}\Vert_2^2 \nonumber \\
    &=\sum_{j=1}^{d}\mathbb{E}\big[[\bv]_j^{2}\big]-\norm{\bu}_2^{2}.
\end{align}
To find $\mathbb{E}[[\bv]_j^2]$, we write $[\bv]_j=\tfrac{1}{m}\sum_{i=1}^{m}[\bc^{(\omega_i)}]_j\langle\bc^{(\omega_i)},\bz\rangle$.
In $\mathbb{E}_{\bc}[[\bc^{(\omega_i)}]_j[\bc^{(\omega_{i'})}]_{j}\langle\bc^{(\omega_i)}, \bz\rangle\langle\bc^{(\omega_{i'})}, \bz\rangle]$, writing $\ell\coloneqq \omega_i-\omega_{i'}$ for the lag between the two sketch indices, there are two cases:
\begin{align}\label{eq:sk_wick}
    &\mathbb{E}_{\bc}[[\bc^{(\omega_i)}]_j[\bc^{(\omega_{i'})}]_{j}\langle\bc^{(\omega_{i})}, \bz\rangle\langle\bc^{(\omega_{i'})}, \bz\rangle] \\
    &~~~~~=\begin{cases}
        \norm{\bu}^{2}+2[\bz]_j^{2}, & i=i', \\
        [\bz]_j^{2}+[\bz]_{j+\ell}[\bz]_{j-\ell}, & o.w.
    \end{cases}\nonumber
\end{align}
whose last term pairs two coordinates at lag $2\ell$ and hence vanishes under $\mathbb{E}_{\bd}$ by \eqref{eq:be_dmean} unless $2\ell\bmod\bar d=0$.
Since $\bar d$ is a power of two, the unique nonzero lag with $2\ell\bmod\bar d=0$ is $\ell^\star=\bar d/2$.
Setting aside the pairs at this lag and summing over $j\le d$ and over the $m$ diagonal and $m(m-1)$ off-diagonal pairs,
\begin{align}\label{eq:sk_sum}
    \sum_{j=1}^{d}\mathbb{E}\big[[\bv]_j^{2}\big]
    &=\frac{m\big(d\norm{\bu}^{2}+2\norm{\bu}^{2}\big)+m(m-1)\norm{\bu}^{2}}{m^{2}}+R\nonumber\\
    &=\frac{(d+1)\norm{\bu}^{2}}{m}+\norm{\bu}^{2}+R,
\end{align}
where $R\ge0$ collects the contributions of that lag.
We bound $R$ as follows.
Let $N\coloneqq|\{(i,i')\in[m]^{2}:i\ne i',\,(\omega_i-\omega_{i'})\bmod\bar d=\ell^\star\}|$ count the ordered sample pairs at lag $\ell^\star$.
For such a pair, $j+\ell^\star\equiv j-\ell^\star\pmod{\bar d}$, so the last term of \eqref{eq:sk_wick} is the \emph{square} $[\bz]_{j+\ell^\star}^{2}=[\bJ\bu]_{j+\ell^\star}^{2}$, which is non-negative and free of $\bd$; hence, with the deterministic quantity $S\coloneqq\sum_{j=1}^{d}[\bJ\bu]_{j+\ell^\star}^{2}$ and $N$ depending only on $\Omega$,
\begin{equation}\label{eq:R_def}
    R=\frac{\mathbb{E}_{\Omega}[N]\,S}{m^{2}},
    \qquad
    0\le S\le\Vert\bJ\bu\Vert_2^{2}=\norm{\bu}_2^{2},
\end{equation}
where the bound on $S$ holds because the $d$ indices $\{j+\ell^\star\}_{j\in[d]}$ are distinct modulo $\bar d$.
Moreover, exactly $\bar d$ ordered pairs $(a,b)\in[\bar d]^{2}$ satisfy $(a-b)\bmod\bar d=\ell^\star$, and each of them lies in $\Omega\times\Omega$ with probability $\frac{m(m-1)}{\bar d(\bar d-1)}$ because $\Omega$ is drawn uniformly without replacement, so that
\begin{equation}\label{eq:R_count}
    \mathbb{E}_{\Omega}[N]=\bar d\cdot\frac{m(m-1)}{\bar d(\bar d-1)}=\frac{m(m-1)}{\bar d-1}.
\end{equation}
Substituting \eqref{eq:R_count} and $S\le\norm{\bu}^{2}$ into \eqref{eq:R_def} gives
\begin{equation}\label{eq:R}
    0\le R\le\frac{(m-1)\norm{\bu}^{2}}{m(\bar d-1)}\le\frac{\norm{\bu}^{2}}{\bar d-1},
\end{equation}
and \eqref{eq:sk_err} gives
\begin{equation}
    \mathbb{E}\big\Vert(\bPhi_t^\top\bPhi_t-\bI_d)\bu\big\Vert^{2}
    \le\Big(\frac{d+1}{m}+\frac{1}{\bar d-1}\Big)\norm{\bu}^{2},
\end{equation}
which completes the second claim.

\paragraph*{Claim (iii)}
Since $\bn$ is independent of $\bPhi_t$ with covariance $N_0\bI_m$, the $j$-th entry of the decoded noise, $[\bPhi_t^\top\bn]_j=\frac{[\bd]_j}{\sqrt m}\sum_{i\in[m]}[\bc^{(\omega_{i})}]_j [\bn]_i$, has
\begin{equation}\label{eq:sk_noise_j}
    \mathbb{E}\Big[\big[\bPhi_t^\top\bn\big]_j^{2}\Big]
    \hspace{-3pt}=\hspace{-8pt}\sum_{i,i'\in[m]}\frac{\mathbb{E}\big[[\bc^{(\omega_{i})}]_j[\bc^{(\omega_{i'})}]_j\big]\mathbb{E}\big[[\bn]_i[\bn]_{i'}\big]}{m}
    =N_0,\nonumber
\end{equation}
identically in $j$, so that summing the $d$ retained entries gives
\begin{equation}\label{eq:sk_claim4}
    \mathbb{E}\left[\big\Vert\bPhi_t^\top\bn\big\Vert^{2}\right]=d\,N_0,
\end{equation}
which completes the third claim.

\section{Proof of \cref{thm:bias}}
\label{app:bias}

We keep the notation of Appendix~\ref{app:be} and $f(x)=\clip_\tau(x)-\alpha_\gamma x$, so that $\bb=\mathbb{E}[\bPhi_t^\top f(\bar{\bs})]/\alpha_\gamma$ by \eqref{eq:bb_kt}.
The proof rests on the observation that, by \eqref{eq:be_inner}, the clipped block is a function of the Gaussian generator $\bc$ alone once $(\bD,\bP_\Omega)$ are fixed.
From $\bPhi_t^\top=\frac{1}{\sqrt m}\bJ^\top\bD\bC^\top\bP_\Omega^\top$ and $[\bC^\top]_{ji}=[\bC]_{ij}=[\bc^{(i)}]_j$, the $j$-th entry of the decoder output is
\begin{equation}\label{eq:bias_entry}
    \big[\bPhi_t^\top f(\bar\bs)\big]_j=\frac{[\bd]_j}{\sqrt m}\sum_{i\in[m]}[\bc^{(\omega_{i})}]_j\,f\big([\bar\bs]_i\big),
    ~~~ j\in[d],
\end{equation}
so that it suffices to show that every term of \eqref{eq:bias_entry} has zero mean over $\bc$.

\paragraph*{Stein's identity}
For $i\in[m]$ and $j\in[d]$, from \eqref{eq:be_inner}, we have 
\begin{equation}\label{eq:bias_coef}
    \frac{\partial[\bar\bs]_i}{\partial [\bc^{(\omega_{i})}]_j}=\frac{1}{\sqrt{m}}\frac{\partial \langle\bc^{(\omega_{i})},\bz\rangle}{\partial [\bc^{(\omega_{i})}]_j}=\frac{[\bz]_j}{\sqrt m}.
\end{equation}
The map $x\mapsto f(x)$ is Lipschitz with constant at most $1+\alpha_\gamma\le2$; therefore $\bc\mapsto f([\bar\bs]_i)$ is Lipschitz and Stein's identity applies, giving
\begin{equation}\label{eq:bias_stein}
    \mathbb{E}_{\bc}\Big[[\bc^{(\omega_{i})}]_j\,f\big([\bar\bs]_i\big)\Big]
    =\frac{[\bz]_j}{\sqrt m}\,\mathbb{E}_{\bc}\Big[\frac{\partial f\big([\bar\bs]_i\big)}{\partial [\bar\bs]_i}\Big].
\end{equation}
Differentiating $f(x)$, we have $\partial f(x)/\partial x=\mathbbm{1}\{|x|<\tau\}-\alpha_\gamma$ for almost every $x$, whence
\begin{equation}\label{eq:bias_zero}
    \mathbb{E}_{\bc}\Big[\frac{\partial f\big([\bar\bs]_i\big)}{\partial [\bar\bs]_i}\Big]
    =\mathbb{P}\big(|[\bar\bs]_i|<\tau\big)-\alpha_\gamma
    \overset{(a)}{=}\mathbb{P}\big(|Z|<\gamma\big)-\alpha_{\gamma}
    =0,
\end{equation}
where $Z\sim\mathcal{N}(0,1)$ and the equality (a) holds because \cref{lem:be} makes $[\bar\bs]_i$ \emph{exactly} $\mathcal{N}(0,\sigma_s^{2})$, while $\tau=\gamma\sigma_s$, \ie, at exactly $\gamma$ standard deviations; the last equality is the definition \eqref{eq:alpha_buss} of the Bussgang gain.
Substituting \eqref{eq:bias_stein} and \eqref{eq:bias_zero} into \eqref{eq:bias_entry}, we have
\begin{equation}\label{eq:bias_assembly}
    \mathbb{E}_{\bc}\Big[\big[\bPhi_t^\top f(\bar\bs)\big]_j\Big]
    =\frac{[\bd]_j}{\sqrt m}\sum_{i\in[m]}\frac{[\bz]_j}{\sqrt m}\underbrace{\mathbb{E}_{\bc}\Big[\frac{\partial f\big([\bar\bs]_i\big)}{\partial[\bar\bs]_i}\Big]}_{=0}
    =0,
\end{equation}
for every $j\in[d]$ and every realization of $(\bD,\bP_\Omega)$; averaging over $(\bd,\Omega)$ and dividing by $\alpha_\gamma>0$ gives $\bb_{k,t}=\mathbf{0}$.

\section{Proof of \cref{prop:moments}}
\label{app:mse}

\paragraph*{Compression loss}
The term $\bee_{\mathrm{sk}}$ can be obtained by reusing \cref{lem:sketch}(ii), because $\bPhi_t$ is common to all EDs and therefore acts on the average update:
\begin{equation}\label{eq:mse_sk}
    \Esk\le \left(\frac{d+1}{m}+\frac{1}{\bar{d}-1}\right) \mathbb{E}\left[\left\Vert\frac{1}{K}\sum_{k}\Delta\bw_{k,t}\right\Vert^{2}\right].
\end{equation}

\paragraph*{Clipping loss}
Because $\Vert\sum_{k\in\mathcal{K}}\ba_k\Vert^{2}\le K\sum_{k\in\mathcal{K}}\Vert\ba_k\Vert^{2}$, and since the $K$ residuals share the sketch $\bPhi_t$ and therefore need not decorrelate across EDs,
\begin{align}\label{eq:mse_cl0}
    \Ecl&=\frac{1}{K^{2}\alpha_\gamma^{2}}\mathbb{E}\left[\left\Vert\sum_{k\in\mathcal{K}}\bPhi_t^\top f_{k,t}(\bar\bs_{k,t})\right\Vert^{2}\right]
    \\
    &\le\frac{1}{\alpha_\gamma^{2}}\cdot\frac{1}{K}\sum_{k\in\mathcal{K}}\mathbb{E}\left[\big\Vert\bPhi_t^\top f_{k,t}(\bar\bs_{k,t})\big\Vert^{2}\right],\nonumber
\end{align}
so it suffices to evaluate one ED, whose indices we drop.
No spectral norm of $\bPhi_t$ is involved; we proceed exactly as in \cref{thm:bias}.
By \eqref{eq:bias_entry} and $|[\bd]_j|=1$, we have
\begin{equation}\label{eq:mse_entry}
    \big\Vert\bPhi_t^\top f(\bar\bs)\big\Vert^{2}=\frac{1}{m}\sum_{j=1}^{d}W_j^{2},
\end{equation}
where $W_j=\sum_{i\in[m]}[\bc^{(\omega_{i})}]_j f\big([\bar\bs]_i\big)$, with $\mathbb{E}[W_j]=0$ by \eqref{eq:bias_assembly}.
In this proof, all vector indices are understood modulo $\bar d$, \ie, $[\bz]_p$ means $[\bz]_{((p-1)\bmod\bar d)+1}$, so that indices exceeding $\bar d$ or falling below $1$ wrap around cyclically.
For fixed $j$ and $\ell=\omega_{i}-\omega_{i'}$, Stein's identity gives us 
\begin{align}\label{eq:ibp2}
    &\mathbb{E}\big[[\bc^{(\omega_{i})}]_j[\bc^{(\omega_{i'})}]_j f([\bar\bs]_i)f([\bar\bs]_{i'})\big] =\mathbbm{1}_{\{i=i'\}}\,\mathbb{E}\big[f([\bar\bs]_i)f([\bar\bs]_{i'})\big]
    \nonumber \\
    &+\frac{1}{m}\mathbb{E}\Big[\frac{\partial^2 f([\bar\bs]_i)}{\partial [\bar\bs]_i^2}f([\bar\bs]_{i'})[\bz]_j[\bz]_{j+\ell}+f([\bar\bs]_i)\frac{\partial^2 f([\bar\bs]_{i'})}{\partial [\bar\bs]_{i'}^2}[\bz]_j[\bz]_{j-\ell}\nonumber\\
    &+\frac{\partial f([\bar\bs]_i)}{\partial [\bar\bs]_i}\frac{\partial f([\bar\bs]_{i'})}{\partial [\bar\bs]_{i'}}\big([\bz]_j^{2}+[\bz]_{j-\ell}[\bz]_{j+\ell}\big)\Big], 
\end{align}
where $\partial^{2}f(x)/\partial x^{2}=\delta_{-\tau}(x)-\delta_{\tau}(x)$ with $\delta_a$ the Dirac delta at $a$, and the identity holds by mollification, since $f$ is Lipschitz and piecewise affine.
The factor $1/m$ and the coefficients $[\bz]_j,[\bz]_{j\pm\ell}$ come from \eqref{eq:bias_coef}.

\emph{For $i=i'$,} we have $\ell=0$, so \eqref{eq:ibp2} reduces to
\begin{equation}\label{eq:mse_diag0}
    \mathbb{E}\big[f([\bar\bs]_i)^{2}\big]
    +\mathbb{E}_{\bc,\Omega}\left[\frac{2[\bz]_j^{2}}{m}\mathbb{E}\bigg[\frac{\partial^2 f([\bar\bs]_i)}{\partial [\bar\bs]_i^2}f([\bar\bs]_i)+\bigg(\frac{\partial f([\bar\bs]_i)}{\partial [\bar\bs]_i}\bigg)^{2}\bigg]\right].
\end{equation}
The first term is exactly $A(\gamma)\sigma_s^{2}$ by \cref{lem:bussgang}.
For the second, we have $\mathbb{E}[(\partial f/\partial x)^{2}]=\mathbb{E}[(\mathbbm{1}_{\{|x|<\tau\}}-\alpha_\gamma)^2]=\alpha_\gamma(1-\alpha_\gamma)$, while $f(\pm\tau)=\pm(1-\alpha_\gamma)\tau$ gives $\mathbb{E}[(\partial^{2}f/\partial x^{2})f]=-2(1-\alpha_\gamma)\gamma\phi(\gamma)$; therefore
\begin{equation}\label{eq:mse_E1}
    \mathbb{E}\bigg[\frac{\partial^{2}f([\bar\bs]_i)}{\partial [\bar\bs]_i^{2}}f([\bar\bs]_i)+\bigg(\frac{\partial f([\bar\bs]_i)}{\partial [\bar\bs]_i}\bigg)^{2}\bigg]\\
    =(1-\alpha_\gamma)\big(\alpha_\gamma-2\gamma\phi(\gamma)\big).
\end{equation}
Summing over the $m$ sampled coordinates and the $d$ retained entries, with $\sum_{j\le d}[\bz]_j^{2}=\norm{\bu}^{2}$ and $\sigma_s^{2}=\norm{\bu}^{2}/m$, the diagonal terms of $\sum_{j\le d}\mathbb{E}[W_j^{2}]$ sum to
\begin{equation}\label{eq:mse_diag}
    d\,A(\gamma)\norm{\bu}^{2}+2\norm{\bu}^{2}(1-\alpha_\gamma)\big(\alpha_\gamma-2\gamma\phi(\gamma)\big).
\end{equation}
\emph{For $i\ne i'$,} the indicator in \eqref{eq:ibp2} is absent, and the two terms carrying a shifted $[\bz]_j$ do \emph{not} have zero mean over the signs.
Conditioned on $(\bd,\Omega)$, the pair $([\bar\bs]_i,[\bar\bs]_{i'})$ is centered Gaussian with variance $\sigma_s^{2}$ and correlation
\begin{equation}\label{eq:mse_corr}
    r_{ii'}=\frac{\mathbb{E}_{\bc}\big[[\bar\bs]_i[\bar\bs]_{i'}\big]}{\sigma_s^{2}}
    =\frac{1}{\norm{\bz}^{2}}\sum_{p=1}^{\bar d}[\bz]_p[\bz]_{p-\ell},
\end{equation}
where every Gaussian factor in \eqref{eq:ibp2} is itself a function of $\bd$, built from the same products $[\bz]_p[\bz]_{p-\ell}$ that appear as its coefficients; the sign average of a coefficient and that of its Gaussian factor cannot be taken separately.
Because $\bJ$ zero-pads, $[\bz]_j=0$ for $j>d$, so summing a coefficient over the retained entries recovers the full circular sum, and
\begin{align}\label{eq:mse_zsum}
    \begin{cases}
    \sum_{j\le d}[\bz]_j[\bz]_{j\pm\ell}=\norm{\bu}_2^{2}\,r_{ii'},\\
    \bigg|\sum_{j\le d}\big([\bz]_j^{2}+[\bz]_{j-\ell}[\bz]_{j+\ell}\big)\bigg|\le2\norm{\bu}_2^{2},
    \end{cases}
\end{align}
where the inequality is obtained by the Cauchy-Schwarz inequality.
Moreover, $\partial^{2}f/\partial x^{2}=\delta_{-\tau}-\delta_{\tau}$ and the density $\phi(\gamma)/\sigma_s$ of $[\bar\bs]_i$ at $\pm\tau$ give
\begin{align}\label{eq:mse_kappa}
    \bigg|\mathbb{E}_{\bc}\bigg[\frac{\partial^{2}f([\bar\bs]_i)}{\partial[\bar\bs]_i^{2}}f([\bar\bs]_{i'})\bigg]\bigg|
    &=\frac{2\phi(\gamma)}{\sigma_s}\Big|\mathbb{E}\big[f([\bar\bs]_{i'})\,\big|\,[\bar\bs]_i=\tau\big]\Big|\nonumber\\
    &\le2\gamma\phi(\gamma)\,\big|r_{ii'}\big|,
\end{align}
where the inequality holds since $|f([\bar\bs]_{i'})|\le \tau=\sigma_s\gamma$.

For the remaining factor, let $H_k$ denote the probabilists' Hermite polynomials, normalized to be orthonormal under the standard Gaussian; their key property is that for a jointly Gaussian pair $(X,X')$ of unit variance and correlation $r$, $\mathbb{E}[H_k(X)H_{k'}(X')]=\mathbbm{1}_{\{k=k'\}}r^{k}$.
Expanding $\partial f(\sigma_sx)/\partial x=\sum_kc_kH_k(x)$, the coefficient $c_0=\mathbb{E}[\partial f/\partial x]$ vanishes by \eqref{eq:bias_zero}, and all odd $c_k$ vanish because $\partial f/\partial x$ is even; hence, with $\sum_kc_k^{2}=\mathbb{E}[(\partial f/\partial x)^{2}]=\alpha_\gamma(1-\alpha_\gamma)$ by Parseval,
\begin{align}\label{eq:mse_hermite}
    \bigg|\mathbb{E}_\bc\bigg[\frac{\partial f([\bar\bs]_i)}{\partial [\bar\bs]_i}\frac{\partial f([\bar\bs]_{i'})}{\partial [\bar\bs]_{i'}}\bigg]\bigg|
    &=\bigg|\sum_{k=2,4,6,\ldots}c_k^{2}\,r_{ii'}^{k}\bigg|\nonumber\\
    &\le r_{ii'}^{2}\,\alpha_\gamma(1-\alpha_\gamma).
\end{align}
Summing the off-diagonal part of \eqref{eq:ibp2} over $j\le d$ and $i\ne i'$, and writing $C_\gamma\coloneqq4\gamma\phi(\gamma)+2\alpha_\gamma(1-\alpha_\gamma)$,
\begin{align}\label{eq:mse_offdiag}
    &\sum_{j\in[d]}\sum_{i\in[m]}\sum_{i'\in[m]\setminus \{i\}}\mathbb{E}_{\bd,\Omega}\bigg[\frac{1}{m}\bigg(\frac{\partial^2 f([\bar\bs]_i)}{\partial [\bar\bs]_i^2}f([\bar\bs]_{i'})[\bz]_j[\bz]_{j+\ell}\nonumber\\
    &\hspace{20pt}+f([\bar\bs]_i)\frac{\partial^2 f([\bar\bs]_{i'})}{\partial [\bar\bs]_{i'}^2}[\bz]_j[\bz]_{j-\ell}\nonumber\\
    &\hspace{20pt}+\frac{\partial f([\bar\bs]_i)}{\partial [\bar\bs]_i}\frac{\partial f([\bar\bs]_{i'})}{\partial [\bar\bs]_{i'}}\big([\bz]_j^{2}+[\bz]_{j-\ell}[\bz]_{j+\ell}\big)\bigg)\bigg]\nonumber\\
    &\le \frac{\norm{\bu}_2^{2}}{m}\mathbb{E}_{\bd,\Omega}\bigg[\sum_{i\in[m]}\sum_{i'\ne i}\bigg(2\bigg|\mathbb{E}_\bc\bigg[\frac{\partial^2 f([\bar\bs]_i)}{\partial [\bar\bs]_i^2}f([\bar\bs]_{i'})\bigg]\bigg|\,\big|r_{ii'}\big|\nonumber\\
    &\hspace{60pt}+2\bigg|\mathbb{E}_\bc\bigg[\frac{\partial f([\bar\bs]_i)}{\partial [\bar\bs]_i}\frac{\partial f([\bar\bs]_{i'})}{\partial [\bar\bs]_{i'}}\bigg]\bigg|\bigg)\bigg]\nonumber\\
    &\overset{(a)}{\le}\frac{C_\gamma\norm{\bu}_2^{2}}{m}\,\mathbb{E}_{\bd,\Omega}\bigg[\sum_{i\in[m]}\sum_{i'\ne i}r_{ii'}^{2}\bigg]
    \overset{(b)}{\le}\frac{C_\gamma\norm{\bu}_2^{2}}{m}\cdot\frac{2m^{2}}{\bar d-1},
\end{align}
where the inequality (a) applies \eqref{eq:mse_kappa} and \eqref{eq:mse_hermite}, both of which are quadratic in $r_{ii'}$ once multiplied by their coefficients.
The inequality (b) holds because $r_{ii'}$ depends on $(i,i')$ only through $\ell=\omega_i-\omega_{i'}\bmod\bar d$, and for a uniform $m$-subset $\Omega$ every ordered pair of distinct elements of $[\bar d]$ appears among $\{(\omega_i,\omega_{i'})\}_{i\ne i'}$ with probability $\frac{m(m-1)}{\bar d(\bar d-1)}$, so that
\begin{equation}\label{eq:be_dsum}
    \mathbb{E}_{\bd,\Omega}\bigg[\sum_{i=1}^{m}\sum_{i'\ne i}r_{ii'}^{2}\bigg]
    =\frac{m(m-1)}{\bar d-1}\sum_{\ell=1}^{\bar d-1}\mathbb{E}_{\bd}\big[r_\ell^{2}\big]
    \le\frac{2m^{2}}{\bar d-1},
\end{equation}
using $\mathbb{E}_{\bd}[r_\ell^{2}]=\frac{1+\mathbbm{1}_{\{2\ell\bmod\bar d=0\}}}{\norm{\bz}^{4}}\sum_{p=1}^{\bar d}[\bz]_p^{2}[\bz]_{p-\ell}^{2}$ and $\sum_{\ell=1}^{\bar d-1}\sum_p[\bz]_p^{2}[\bz]_{p-\ell}^{2}=\norm{\bz}^{4}-\norm{\bz}_4^{4}\le\norm{\bz}^{4}$.
Adding \eqref{eq:mse_diag} and \eqref{eq:mse_offdiag}, dividing by $m$ as in \eqref{eq:mse_entry}, and factoring out the leading term, we have
\begin{equation}\label{eq:mse_cl}
    \mathbb{E}[\Vert\bPhi_t^\top f(\bar\bs)\Vert^{2}]\le\frac{d}{m}A(\gamma)\,\norm{\bu}^{2}\big(1+\varepsilon_{\mathrm{cl}}\big),
\end{equation}
where
\begin{align}\label{eq:mse_epsd}
    \varepsilon_{\mathrm{cl}}=\frac{1}{d\,A(\gamma)}\Big[&2(1-\alpha_\gamma)\big(\alpha_\gamma-2\gamma\phi(\gamma)\big)\nonumber\\
    &+\big(4\alpha_\gamma(1-\alpha_\gamma)+8\gamma\phi(\gamma)\big)\tfrac{m}{\bar d-1}\Big].
\end{align}

\paragraph*{Channel noise}
By substituting $c_t$ by $c_t^\star$ in \eqref{eq:cstar}, we have the following identity from \cref{lem:sketch}(iii):
\begin{align}\label{eq:mse_ch}
    \Ech&=\frac{\mathbb{E}[\Vert\bPhi_t^\top\bn_t\Vert^{2}]}{c_t^{2}K^{2}}= \frac{dN_0}{K^2P_\mathrm{pk}}\cdot \mathbb{E}\left[\max_{k\in\mathcal{K}}\frac{\min\{\tau_{k,t}^2,\Vert\bar{\bs}\Vert_\infty^2\}}{h_{k,t}^2\alpha_\gamma^2}\right]\nonumber\\ 
    & \le \frac{dN_0}{\alpha_\gamma^2KP_\mathrm{pk}} \cdot \max_{k\in\mathcal{K}}\frac{\gamma^2\Vert\Delta\bw_{k,t}\Vert_2^2}{mKh_{k,t}^2} \\ 
    & \le \frac{d\gamma^2N_0}{m\alpha_\gamma^2KP_\mathrm{pk}} \mathbb{E}\left[ \max_{k\in\mathcal{K}} h_{k,t}^{-2} \right]\cdot \max_{k\in\mathcal{K}}\frac{\mathbb{E}[\Vert\Delta\bw_{k,t}\Vert_2^2]}{K}. \nonumber
\end{align}
The noise $\bn_t$ is zero-mean and independent of $(\bd,\bg,\Omega)$ and of the local updates; hence $\bee_{\mathrm{ch}}$ is uncorrelated with $\bee_{\mathrm{sk}}+\bee_{\mathrm{cl}}$.
For the remaining two terms, the Cauchy--Schwarz inequality gives $\mathbb{E}\Vert\bee_{\mathrm{sk}}+\bee_{\mathrm{cl}}\Vert^{2}\le(\sqrt{\Esk}+\sqrt{\Ecl})^{2}$, and adding \eqref{eq:mse_ch} completes \eqref{eq:moments}.

\section{Proof of \cref{thm:conv}}
\label{app:conv}

Here, we prove the convergence of the proposed method.
For notational brevity, we define  the average local iterate $\bar\bw_{G,t,i}=\frac{1}{K}\sum_{k\in\mathcal{K}}\bw_{k,t,i}$, $\bar\bw_{G,t,0}=\bw_{G,t}$,  and $\Delta\bw_{G,t} = \bar\bw_{G,t,I}-\bw_{G,t}$.
At the end of round $t$ the EDs update the global model by $\bw_{G,t+1}=\bw_{G,t}+\Delta\widehat{\bw}_{G,t}$, and since $\bar\bw_{G,t,I}=\bw_{G,t}+\Delta\bw_{G,t}$, we may split the round into the part contributed by the transceiver and the part contributed by the local steps:
\begin{align}\label{eq:conv_split}
    &\mathbb{E}\big[F(\bw_{G,t+1})-F(\bw_{G,t})\big] = \underbrace{\mathbb{E}\big[F(\bw_{G,t+1})-F(\bar\bw_{G,t,I})\big]}_{\textbf{Part A}:\ \text{transceiver error of \AirGCCD}}\nonumber\\
    & ~~~ +\sum_{i\in[I]}\underbrace{\mathbb{E}\big[F(\bar\bw_{G,t,i})-F(\bar\bw_{G,t,i-1})\big]}_{\textbf{Part B}:\ \text{local updates}}.
\end{align}

\paragraph*{Part A}
From the assumption A2 in Assumption \ref{assumption:1} and $\bw_{G,t+1}-\bar\bw_{G,t,I}=\Delta\widehat{\bw}_{G,t}-\Delta\bw_{G,t}$, we have
\begin{align}\label{eq:conv_A_smooth}
    &\mathbb{E}\big[F(\bw_{G,t+1})-F(\bar\bw_{G,t,I})\big] ~~\le \underbrace{\frac{\beta}{2}\mathbb{E}\Big[\big\Vert\Delta\widehat{\bw}_{G,t}-\Delta\bw_{G,t}\big\Vert^{2}\Big]}_{\textbf{Part A1}} \nonumber\\
    &~~~~~+\underbrace{\mathbb{E}\Big[\nabla F(\bar\bw_{G,t,I})^\top\big(\Delta\widehat{\bw}_{G,t}-\Delta\bw_{G,t}\big)\Big]}_{\textbf{Part A2}}.
\end{align}
\paragraph*{Part A1}\cref{prop:moments} gives $\mathbb{E}\Vert\Delta\widehat{\bw}_{G,t}-\Delta\bw_{G,t}\Vert^{2}\le2\Esk+2\Ecl+\Ech$, and it remains to express the three terms through the local gradients.
Using $\Vert\sum_{i\in[I]}\ba_i\Vert^{2}\le I\sum_{i\in[I]}\Vert\ba_i\Vert^{2}$, together with the assumption A4 in Assumption \ref{assumption:1}, we have
\begin{align}\label{eq:conv_norms}
    \begin{cases}
    \max_k\mathbb{E}[\frac{\norm{\Delta\bw_{k,t}}^2_2}{K}] \le  
    \sum_{k}\frac{\mathbb{E}\big[\norm{\Delta\bw_{k,t}}_2^{2}\big]}{K} \le\eta^{2}I^{2}G_2^{2} \\ 
    \mathbb{E}[\Vert\Delta\bw_{G,t}\Vert^{2}_2]\le  
    \sum_{k}\frac{\mathbb{E}\big[\norm{\Delta\bw_{k,t}}_2^{2}\big]}{K}\le\eta^{2}I^{2}G_2^{2}.
    \end{cases}
\end{align}
Substituting \eqref{eq:conv_norms}  into \eqref{eq:esk_cl_ch} and recalling the constants \eqref{eq:deltadef}, we finally have
\begin{equation}\label{eq:conv_A2}
    (\textbf{Part A1})\le\frac{\beta\eta^{2}I^{2}G_2^{2}}{2}\big(2\Delta_{\mathrm{cl}}+\Delta_{\mathrm{ch}}\big) + \beta\Delta_{\mathrm{sk}}\Vert\Delta\bw_{G,t}\Vert^2.
\end{equation}

\paragraph*{Part A2} 
The sketch $(\bD,\bC,\bP_\Omega)$ and the noise $\bn_t$ are drawn independently of the local training, so $\bar\bw_{G,t,I}$ is a function of $\bw_{G,t}$ and $\{\Delta\bw_{k,t}\}_{k\in\mathcal{K}}$; conditioning on them and applying \cref{thm:bias},
\begin{equation}\label{eq:conv_A1}
\begin{split}
    &(\textbf{Part A2})=\mathbb{E}\Big[\nabla F(\bar\bw_{G,t,I})^\top\\
    &\quad\times\underbrace{\mathbb{E}\big[\Delta\widehat{\bw}_{G,t}-\Delta\bw_{G,t}\mid\bw_{G,t},\{\Delta\bw_{k,t}\}_{k\in\mathcal{K}}\big]}_{=\mathbf{0}\ \text{by \cref{thm:bias}}}\Big]=0.
\end{split}
\end{equation}

\paragraph*{Part B}
For each $i\in[I]$, we have $\bar\bw_{G,t,i}-\bar\bw_{G,t,i-1}=-\frac{\eta}{K}\sum_{k\in\mathcal{K}}\bg_{k,t,i-1}$, so the assumption A2 and A3 give
\begin{equation}\label{eq:conv_B_smooth}
\begin{split}
    &\mathbb{E}\big[F(\bar\bw_{G,t,i})-F(\bar\bw_{G,t,i-1})\big] \\
    &~~\le-\frac{\eta}{K}\mathbb{E}\Big[\nabla F(\bar\bw_{G,t,i-1})^\top\sum_{k\in\mathcal{K}}\nabla F_k(\bw_{k,t,i-1})\Big] \\
    &~~~~~+\frac{\eta^{2}\beta}{2K^{2}}\mathbb{E}\Big[\Big\Vert\sum_{k\in\mathcal{K}}\bg_{k,t,i-1}\Big\Vert^{2}\Big].
\end{split}
\end{equation}
Because $-\ba^\top\bb=-\frac12\Vert\ba\Vert^{2}-\frac12\Vert\bb\Vert^{2}+\frac12\Vert\ba-\bb\Vert^{2}$ holds, we have
\begin{align}\label{eq:conv_B}
    &(\textbf{Part B})\le-\underbrace{\frac{\eta}{2}\mathbb{E}\big[\Vert\nabla F(\bar\bw_{G,t,i-1})\Vert^{2}\big]}_{\textbf{Part B1}} \\
    &-\underbrace{\frac{\eta}{2}\mathbb{E}\Big[\Big\Vert\frac{1}{K}\sum_{k\in\mathcal{K}}\nabla F_k(\bw_{k,t,i-1})\Big\Vert^{2}\Big]}_{\textbf{Part B2}}\nonumber\\
    &+\underbrace{\frac{\eta}{2}\mathbb{E}\Big[\Big\Vert\nabla F(\bar\bw_{G,t,i-1})-\frac{1}{K}\sum_{k\in\mathcal{K}}\nabla F_k(\bw_{k,t,i-1})\Big\Vert^{2}\Big]}_{\textbf{Part B3}}\nonumber\\
    &+\underbrace{\frac{\eta^{2}\beta}{2K^{2}}\mathbb{E}\Big[\Big\Vert\sum_{k\in\mathcal{K}}\bg_{k,t,i-1}\Big\Vert^{2}\Big]}_{\textbf{Part B4}}.\nonumber
\end{align}

\paragraph*{Parts B2 and B4}
Since the mini-batches are drawn independently across EDs by the assumption A3, the expectation term of \textbf{Part B4} can be written as 
\begin{equation}\label{eq:conv_B4}
\begin{split}
    & \mathbb{E}\Big[\Big\Vert\sum_{k\in\mathcal{K}}\bg_{k,t,i-1}\Big\Vert^{2}\Big]
    =\mathbb{E}\Big[\Big\Vert\sum_{k\in\mathcal{K}}\nabla F_k(\bw_{k,t,i-1})\Big\Vert^{2}\Big] \\
    &~~~~~~~~+\underbrace{\sum_{k\in\mathcal{K}}\mathbb{E}\big[\Vert\bg_{k,t,i-1}-\nabla F_k(\bw_{k,t,i-1})\Vert^{2}\big]}_{\le K\xi^{2}}.
\end{split}
\end{equation}
Summing \textbf{Part B4} minus \textbf{Part B2} over $i\in[I]$ and adding the second term of \textbf{Part A1} in \eqref{eq:conv_A2},  and using $\Vert\sum_{l=1}^{n}\ba_l\Vert^{2}\le n\sum_{l=1}^{n}\Vert\ba_l\Vert^{2}$, their combined coefficient in front of $\sum_{i\in[I]}\mathbb{E}\Vert\frac{1}{K}\sum_{k\in[K]}\nabla F_k(\bw_{k,t,i-1})\Vert^{2}$ is
\begin{equation}\label{eq:conv_B24}
    \frac{\eta^{2}\beta}{2}+\beta\Delta_{\mathrm{sk}}\eta^{2}I-\frac{\eta}{2}
    =\frac{\eta}{2}\Big(\eta\beta\big(1+2I\Delta_{\mathrm{sk}}\big)-1\Big)\ \le\ 0,
\end{equation}
where the inequality is exactly the step-size condition $\eta\le 1/(\beta(1+2I\Delta_\mathrm{sk}))$ of \cref{thm:conv}.
Then, we have 
\begin{align}\label{eq:conv_B24final}
    &\sum_{i\in[I]}\big[-(\textbf{Part B2})+(\textbf{Part B4})\big]+\beta\Delta_{\mathrm{sk}}\mathbb{E}[\Vert\Delta\bw_{G,t}\Vert^2]\\
    &~~~~\le\frac{\eta^{2}\beta I\xi^{2}}{2K}(1+2I\Delta_\mathrm{sk}).\nonumber
\end{align}

\paragraph*{Part B3}
Since $\nabla F=\frac1K\sum_k\nabla F_k$, the assumption A2 and $\Vert\frac{1}{K}\sum_k\ba_k\Vert^{2}\le\frac1K\sum_k\Vert\ba_k\Vert^{2}$ give
\begin{align}\label{eq:conv_B3a}
    &(\textbf{Part B3}) \nonumber\\ 
    &=\frac{\eta}{2}\mathbb{E}\Big[\Big\Vert\frac{1}{K}\sum_{k\in\mathcal{K}}\big(\nabla F_k(\bar\bw_{G,t,i-1})-\nabla F_k(\bw_{k,t,i-1})\big)\Big\Vert^{2}\Big]\nonumber\\
    &\le\frac{\eta\beta^{2}}{2K}\sum_{k\in\mathcal{K}}\mathbb{E}\big\Vert\bar\bw_{G,t,i-1}-\bw_{k,t,i-1}\big\Vert^{2},\nonumber\\
    &=\frac{\eta^{3}\beta^2}{2K}\sum_{k\in\mathcal{K}}\mathbb{E}\Big\Vert\sum_{l=0}^{i-2}\big(\bg_{k,t,l}-\bar\bg_{t,l}\big)\Big\Vert^{2} \\
    &\overset{(a)}{\le}\frac{\eta^{3}\beta^2(i-1)}{2K}\sum_{k\in\mathcal{K}}\sum_{l=0}^{i-2}\mathbb{E}\big\Vert\bg_{k,t,l}-\bar\bg_{t,l}\big\Vert^{2} \nonumber\\
    &\overset{(b)}{\le}\frac{\eta^{3}\beta^2(i-1)}{2}\sum_{l=0}^{i-2}\frac{1}{K}\sum_{k\in\mathcal{K}}\mathbb{E}\big\Vert\bg_{k,t,l}\big\Vert^{2}
    \ \le\ \frac{\eta^{3}\beta^{2}(i-1)^{2}G_2^{2}}{2},\nonumber
\end{align}
where $\bar\bg_{t,l}=\frac{1}{K}\sum_{k\in\mathcal{K}}\bg_{k,t,l}$, the inequality (a) holds because $\Vert\sum_{l=1}^{n}\ba_l\Vert^{2}\le n\sum_{l=1}^{n}\Vert\ba_l\Vert^{2}$, and the inequality (b) holds because $\frac1K\sum_k\Vert\ba_k-\bar\ba\Vert^{2}\le\frac1K\sum_k\Vert\ba_k\Vert^{2}$, the last step using the assumption A4.

\paragraph*{Results}
From \eqref{eq:conv_split}, \eqref{eq:conv_A1}, \eqref{eq:conv_A2}, \eqref{eq:conv_B}, \eqref{eq:conv_B24final}, \eqref{eq:conv_B3a}, and using $\sum_{i\in[I]}(i-1)^{2}\le\frac{(I-1)I^{2}}{3}$, we have
\begin{align}\label{eq:conv_round}
    &\mathbb{E}\big[F(\bw_{G,t+1})-F(\bw_{G,t})\big] \\
    &~~\le-\frac{\eta}{2}\sum_{i\in[I]}\mathbb{E}\big[\Vert\nabla F(\bar\bw_{G,t,i-1})\Vert^{2}\big]
    +\frac{I\eta^{2}\beta\xi^{2}}{2K}(1+2I\Delta_\mathrm{sk}) \nonumber\\
    &~~~~~+\frac{\eta^{3}\beta^{2}(I-1)I^{2}G_2^{2}}{6}
    +\frac{\beta\eta^{2}I^{2}G_2^{2}}{2}\big(2\Delta_{\mathrm{cl}}+\Delta_{\mathrm{ch}}\big).\nonumber
\end{align}
Summing \eqref{eq:conv_round} from $t=0$ to $t=T-1$, the left-hand side is $\mathbb{E}[F(\bw_{G,T})]-F(\bw_{G,0})\ge-\Delta_F$ by the assumption A1; rearranging and dividing by $\eta TI/2$ then gives
\begin{align}\label{eq:conv_final}
    &\frac{1}{TI}\sum_{t=0}^{T-1}\sum_{i\in[I]}\mathbb{E}\big\Vert\nabla F(\bar\bw_{G,t,i-1})\big\Vert^{2}
    \le\frac{2\Delta_F}{\eta TI}+\frac{\eta\beta\xi^{2}}{K}(1+2I\Delta_\mathrm{sk}) \nonumber\\
    &~~~~~+\eta\beta I\big(2\Delta_{\mathrm{cl}}+\Delta_{\mathrm{ch}}\big)G_2^{2}
    +\frac{\eta^{2}\beta^{2}(I-1)IG_2^{2}}{3},
\end{align}
Finally, substituting $\eta=1/\sqrt{TI}$ into \eqref{eq:conv_final}, which meets the step-size condition of \cref{thm:conv} by construction, we obtain
\begin{equation}\label{eq:conv_rate}
\begin{split}
    &\frac{1}{TI}\sum_{t=0}^{T-1}\sum_{i\in[I]}\mathbb{E}\big\Vert\nabla F(\bar\bw_{G,t,i-1})\big\Vert^{2} \\
    &~~\le\frac{1}{\sqrt{TI}}\Big[2\Delta_F+\frac{\beta\xi^{2}}{K}(1+2I\Delta_\mathrm{sk})+\beta I\big(2\Delta_{\mathrm{cl}}+\Delta_{\mathrm{ch}}\big)G_2^{2}\Big] \\
    &~~~~~+\frac{\beta^{2}(I-1)G_2^{2}}{3T},
\end{split}
\end{equation}
which is \eqref{eq:convrate}.

\section{Proof of \cref{cor:opt}}
\label{app:opt}

\paragraph*{Monotonicity of $   \Psi(\cdot)$}

From the definition of $\alpha_\gamma$, $\omega(\gamma)$, and $A(\gamma)$, we have $\frac{\partial\alpha_\gamma}{\partial\gamma}=2\phi(\gamma)$, $\frac{\partial\omega(\gamma)}{\partial\gamma}=4\gamma Q(\gamma)$,  and 
\begin{equation}\label{eq:opt_derivs}
    \frac{\partial A(\gamma)}{\partial\gamma}
    =\frac{\partial\omega(\gamma)}{\partial\gamma}-2\alpha_\gamma\frac{\partial\alpha_\gamma}{\partial\gamma}
    =4\big(\gamma Q(\gamma)-\alpha_\gamma\phi(\gamma)\big).
\end{equation}
Then, we have 
\begin{align}\label{eq:focquot}
    &\frac{\partial J(\gamma;\snr)}{\partial\gamma}
    =\frac{2}{\alpha_\gamma^{3}}\bigg[\alpha_\gamma\Big(4\big(\gamma Q(\gamma)-\alpha_\gamma\phi(\gamma)\big)+\frac{\gamma}{\snr}\Big)  \nonumber \\ 
    &~~~~~~~~~~~~~~~~~~~~~~~~~~~~~ -2\phi(\gamma)\Big(2A(\gamma)+\frac{\gamma^{2}}{\snr}\Big)\bigg].
\end{align}
Using $A(\gamma)+\alpha_\gamma^{2}=\omega(\gamma)$, the terms inside the bracket of \eqref{eq:focquot} are rewritten as 
\begin{equation}\label{eq:focfold}
    \frac{\gamma}{\snr}\big(\alpha_\gamma-2\gamma\phi(\gamma)\big)-4\big(\phi(\gamma)\omega(\gamma)-\gamma\alpha_\gamma Q(\gamma)\big).
\end{equation}
Let $\Gamma(\gamma)=\alpha_\gamma-2\gamma\phi(\gamma)=\mathbb{E}[Z^{2}\mathbbm{1}_{\{|Z|\le\gamma\}}]$, which is positive by \eqref{eq:normal_property}.
Substituting $\omega(\gamma)=\Gamma(\gamma)+2\gamma^{2}Q(\gamma)$ into the second term of \eqref{eq:focfold},
\begin{equation}\label{eq:opt_factor}
    \phi(\gamma)\omega(\gamma)-\gamma\alpha_\gamma Q(\gamma)
    =\Gamma(\gamma)\big(\phi(\gamma)-\gamma Q(\gamma)\big),
\end{equation}
so that \eqref{eq:focfold} equals $\Gamma(\gamma)\big(\gamma/\snr-4(\phi(\gamma)-\gamma Q(\gamma))\big)$ and \eqref{eq:focquot} becomes
\begin{equation}\label{eq:focsign}
    \frac{\partial J(\gamma;\snr)}{\partial\gamma}
    =\frac{8\,\Gamma(\gamma)\big(\phi(\gamma)-\gamma Q(\gamma)\big)}{\alpha_\gamma^{3}}\Big(\frac{\Psi(\gamma)}{\snr}-1\Big),
\end{equation}
with $\Psi(\gamma)=\gamma/\big(4(\phi(\gamma)-\gamma Q(\gamma))\big)$ of \eqref{eq:Psidef}.
By Mills' inequality $Q(\gamma)<\phi(\gamma)/\gamma$, the factor $\phi(\gamma)-\gamma Q(\gamma)$ is positive on $(0,\infty)$; hence, the sign of $\partial J/\partial\gamma$ is that of $\Psi(\gamma)-\snr$.
From $\partial\phi(\gamma)/\partial\gamma=-\gamma\phi(\gamma)$ and $\partial Q(\gamma)/\partial\gamma=-\phi(\gamma)$, we have 
\begin{equation}\label{eq:opt_Psiprime}
    \frac{\partial\Psi(\gamma)}{\partial\gamma}
    =\frac{\phi(\gamma)}{4\big(\phi(\gamma)-\gamma Q(\gamma)\big)^{2}}>0,
\end{equation}
where the equality holds since $\frac{\partial}{\partial\gamma}\big(\phi(\gamma)-\gamma Q(\gamma)\big)=-Q(\gamma)$.
Thus $\Psi$ is a strictly increasing bijection of $(0,\infty)$ onto itself and by \eqref{eq:focsign} $J(\cdot;\snr)$ is strictly decreasing on $(0,\Psi^{-1}(\snr))$ and strictly increasing on $(\Psi^{-1}(\snr),\infty)$.
The minimizer therefore exists, is unique for every $\snr>0$, and is \eqref{eq:mustar}.

\paragraph*{The optimality gap \eqref{eq:gapbound}}
Write $r(\gamma)\coloneqq\frac{2A(\gamma)}{\alpha_\gamma^{2}}\varepsilon_{\mathrm{cl}}$ with $\varepsilon_{\mathrm{cl}}$ of \eqref{eq:mse_epsd}, so that by \eqref{eq:Jsplit} the clipping-and-channel term is $J(\gamma;\snr)+r(\gamma)$; the clipped power $A(\gamma)$ cancels and
\begin{align}\label{eq:opt_r}
    r(\gamma)=\frac{4}{d\,\alpha_\gamma^{2}}\Big[&(1-\alpha_\gamma)\big(\alpha_\gamma-2\gamma\phi(\gamma)\big)\nonumber\\
    &+\big(2\alpha_\gamma(1-\alpha_\gamma)+4\gamma\phi(\gamma)\big)\tfrac{m}{\bar d-1}\Big].
\end{align}
Since $\alpha_\gamma-2\gamma\phi(\gamma)=\mathbb{E}[Z^{2}\mathbbm{1}_{\{|Z|\le\gamma\}}]$ by \eqref{eq:normal_property}, we have $0\le\alpha_\gamma-2\gamma\phi(\gamma)\le\alpha_\gamma$, and $\alpha_\gamma(1-\alpha_\gamma)\le\tfrac14$ for $\alpha_\gamma\in(0,1)$; moreover $\gamma\phi(\gamma)\le\phi(1)<\tfrac14$, since $\frac{\partial}{\partial\gamma}\gamma\phi(\gamma)=(1-\gamma^{2})\phi(\gamma)$ vanishes only at $\gamma=1$, so that $2\alpha_\gamma(1-\alpha_\gamma)+4\gamma\phi(\gamma)\le\tfrac32$; hence
\begin{equation}\label{eq:opt_rbound}
    0\le r(\gamma)\le\frac{1}{d\,\alpha_\gamma^{2}}\Big(1+\frac{6m}{\bar d-1}\Big),~\forall\gamma>0.
\end{equation}
Because $r\ge0$, $\min_\gamma(J+r)\ge\min_\gamma J=J(\gamma^\star;\snr)$, and therefore
\begin{align}\label{eq:opt_gap}
    &\big(J(\gamma^\star;\snr)+r(\gamma^\star)\big)-\min_\gamma\big(J(\gamma;\snr)+r(\gamma)\big)\\
    & \le r(\gamma^\star) 
    \le\frac{1}{d\,\alpha_{\gamma^\star}^{2}}\Big(1+\frac{6m}{\bar d-1}\Big),\nonumber
\end{align}
which completes the claim.

\end{document}